\documentclass{article}

\usepackage{arxiv}

\usepackage[utf8]{inputenc} 
\usepackage[T1]{fontenc}    
\usepackage{hyperref}       
\usepackage{url}            
\usepackage{booktabs}       
\usepackage{amsfonts}       
\usepackage{nicefrac}       
\usepackage{microtype}      
\usepackage{lipsum}
\usepackage{graphicx}
\graphicspath{ {./images/} }

\usepackage{multirow}                 
\usepackage{multicol}    
\usepackage{tabularx}
\usepackage{amsmath}
\usepackage{color,xcolor}
\usepackage{caption}
\usepackage{subcaption}
\usepackage{wrapfig}

\definecolor{darkpurple}{RGB}{48, 0, 63} 

\hypersetup{
	colorlinks=true,
	linkcolor=cyan,
	filecolor=blue,      
	urlcolor=red,
	citecolor=green,
}

\title{Revisiting Attention for Multivariate Time Series Forecasting}

\author{
  Haixiang Wu \\
  School of Computer Science and Communication Engineering, Jiangsu University\\
  Zhenjiang, China 212013 \\
  \texttt{2232208015@stmail.ujs.edu.cn} | \texttt{425059276@qq.com} \\
}

\begin{document}
\maketitle
\begin{abstract}
Current Transformer methods for Multivariate Time-Series Forecasting (MTSF) are all based on the conventional attention mechanism. They involve sequence embedding and performing a linear projection of Q, K, and V, and then computing attention within this latent space. We have never delved into the attention mechanism to explore whether such a mapping space is optimal for MTSF. To investigate this issue, this study first proposes Frequency Spectrum attention (FSatten), a novel attention mechanism based on the frequency domain space. It employs the Fourier transform for embedding and introduces Multi-head Spectrum Scaling (MSS) to replace the conventional linear mapping of Q and K. FSatten can accurately capture the periodic dependencies between sequences and outperform the conventional attention without changing mainstream architectures. We further design a more general method dubbed Scaled Orthogonal attention (SOatten). We propose an orthogonal embedding and a Head-Coupling Convolution (HCC) based on the neighboring similarity bias to guide the model in learning comprehensive dependency patterns. Experiments show that FSatten and SOatten surpass the SOTA which uses conventional attention, making it a good alternative as a basic attention mechanism for MTSF. The codes and log files will be released at: https://github.com/Joeland4/FSatten-SOatten.
\end{abstract}


\section{Introduction}

\begin{wrapfigure}{r}{6cm}
    \vspace{-5mm}
    \centering
    \includegraphics[width=0.35\textwidth]{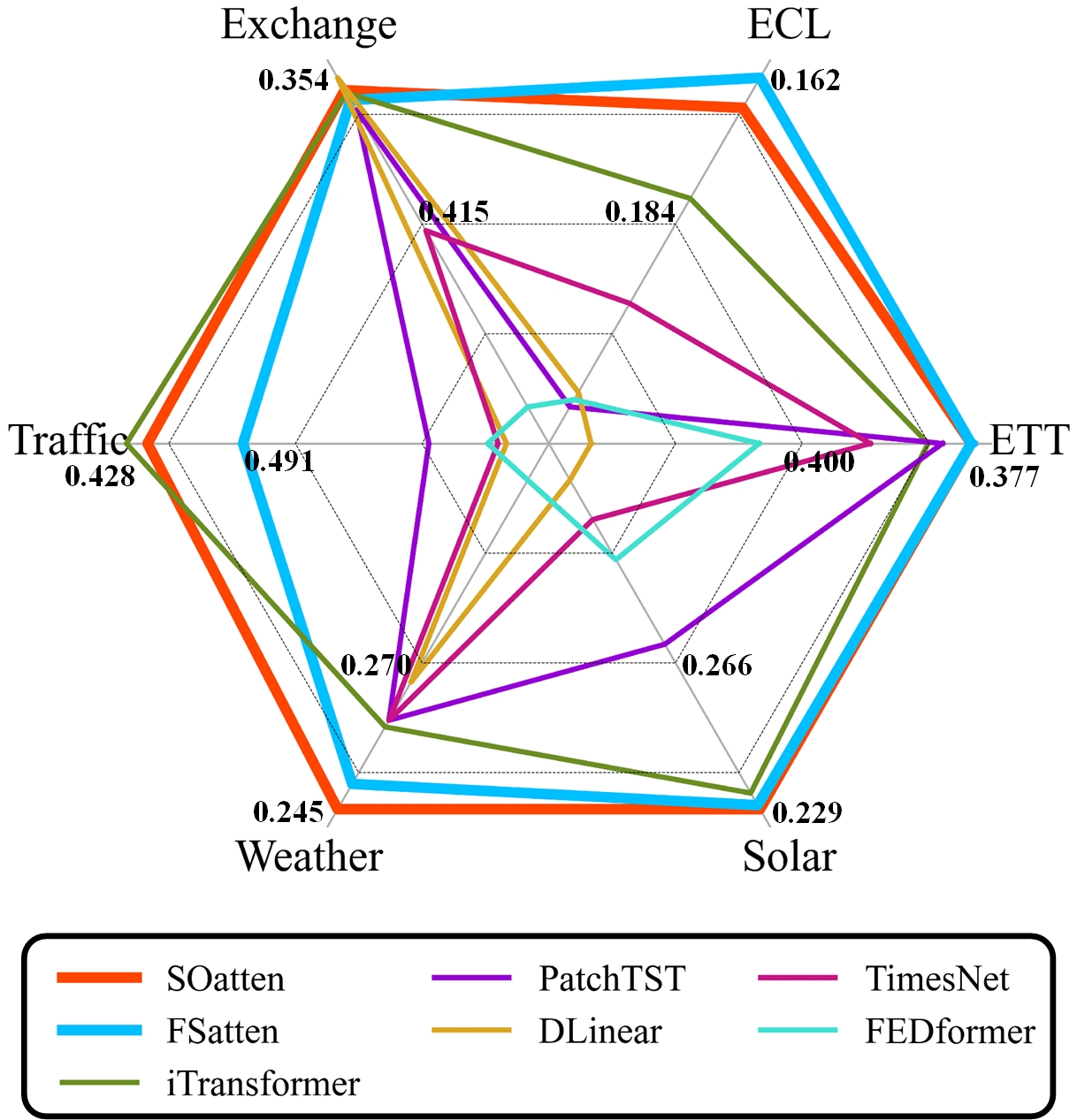}
    \caption{\footnotesize Performance of FSatten and SOatten.}
    \label{radio}
    \vspace{-5mm}
\end{wrapfigure}

Multivariate Time Series Forecasting (MTSF) is extensively applied in real-world scenarios such as finance, electricity, and transportation. Benefiting from the attention mechanism's \cite{vaswani2017attention} ability to effectively capture both long- and short-term dependencies, many Transformer-based methods have demonstrated remarkable performance. These methods mainly include the Temporal Transformer, which evolves from applying attention between time steps \cite{zhou2021informer} \cite{li2019enhancing} \cite{liu2021pyraformer} to applying attention between subseries \cite{wu2021autoformer} \cite{zhou2022fedformer} \cite{nie2022time}, and the Variate Transformer, which explicitly models the correlations between variates though attention \cite{zhang2022crossformer} \cite{liu2023itransformer}.

\begin{figure}
    \centering
    \includegraphics[width=\textwidth]{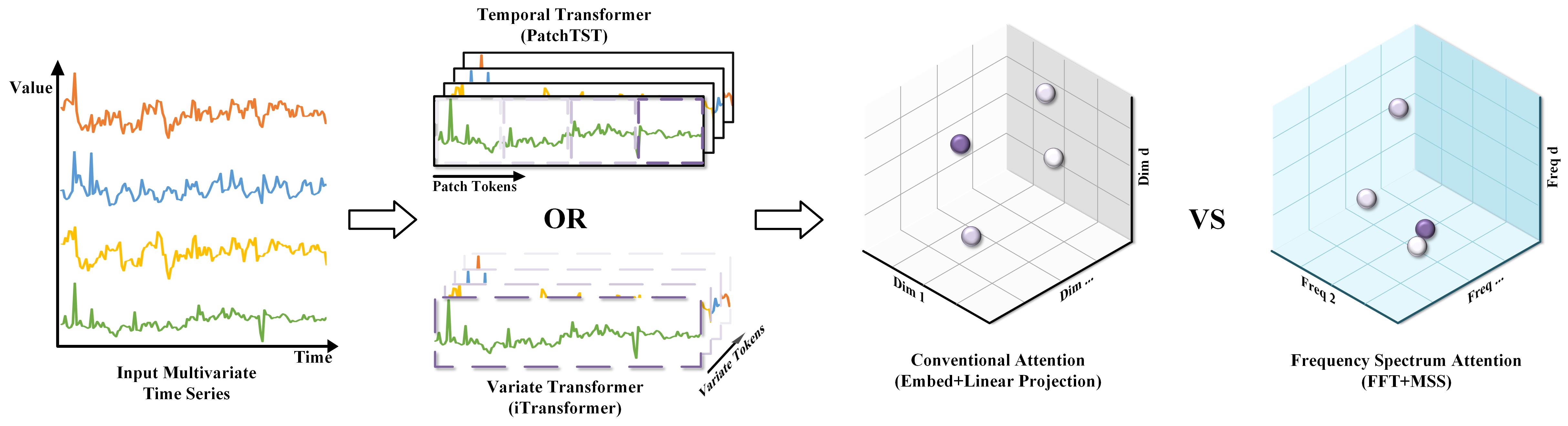}
    \caption{(Left) Temporal Transformer and Variate Transformer. (Right) Comparison of mapping space from conventional attention and FSatten}
    \label{MTSF Transformer}
\end{figure}

Temporal and Variate Transformers, as shown in Figure \ref{MTSF Transformer}, primarily apply attention mechanisms to time series sequences and have become mainstream, with many subsequent studies \cite{zhou2023one} \cite{jin2023time} building upon these architectures. We aim to understand where these correlations between sequences manifest. In the attention mechanism, sequences are mapped to a learnable space by embedding and linear projection, and then the correlations are calculated within this latent space. Although we cannot provide physical interpretations for the learned characteristics of black-box neural networks, it is worth considering whether the dependency capturing within this latent space is optimal.

To explore the interpretability of time-series attention and make further improvements for MTSF, we propose Frequency Spectrum attention (FSatten), which is based on the frequency domain space. The consideration is that dependencies between non-stationary sequences are complex and can be synchronous or asynchronous at different frequencies. It is appropriate to consider these dependencies from the frequency domain perspective, as previous works have made improvements \cite{zhou2022fedformer} \cite{xu2023fits}. In FSatten, Fourier transform is utilized for the embedding, and Query and Key are projected by a proposed Multi-head Spectrum Scaling (MSS) instead of the conventional linear projection. MSS scales amplitude for different frequency components under each of the multiple heads, identifying clear frequency spectral relationships between sequences.

Experimental results in Figure \ref{radio} and Table \ref{Average Results} show that without modifying the architecture, simply replacing the conventional attention with FSatten yields significant improvement over the state-of-the-art (SOTA), especially for scenarios with significant periodicity. This suggests that the conventional attention mechanism is not optimal for MTSF. However, the frequency domain space cannot meet all the characteristics of different scenes, indicating that there is more to explore. Also, FSatten is good for capturing same-frequency correlations between variates but may not be highly appropriate for Temporal Transformers, as sequences split from one variate naturally tend to exhibit the same periodic frequency.

To find a more general method for various scenarios, we propose Scaled Orthogonal attention (SOatten), which creates a learnable orthogonal transformation beyond the Fourier transform. In SOatten, we propose a Head Coupling Convolution (HCC) to guide the updating of learnable orthogonal spaces by leveraging the similarity between adjacent sequences. Experiments show that SOatten enhances overall performance compared to FSatten when applied to Variate Transformer, iTransformer \cite{liu2023itransformer} and makes significant improvements when applied to a general Temporal Transformer, PatchTST \cite{nie2022time}, showcasing stronger adaptability. We hope the proposed methods may inspire future work in time series analysis and offer contributions to other deep learning fields.

The main contributions of this work are as follows:
\begin{itemize}
\item We propose FSatten, a more interpretable and effective model than conventional attention for MTSF, which replaces the learnable latent space by Frequency domain. 
\item Through the proposed MSS mapping for Query and Key, FSatten accurately identifies the frequency correlations between sequences. This specific dependency is more effective than what is provided by conventional linear projections. 
\item We propose SOatten, a more general attention than FSatten which provides a learnable orthogonal latent space facilitated by a designed HCC module for capturing comprehensive dependencies.
\item On six real-world long-term forecasting benchmarks, our FSatten and SOatten outperform the SOTA method which utilizes conventional attention by an overall of 8.1\% and 21.8\% on MSE, demonstrating their superior effectiveness for MTSF.
\end{itemize}

\section{Preliminaries}
A Multivariate Time Series (MTS) sampling with look back window $L$ is denoted by $X\ = \left\{ x_1,...,x_L \right\} \epsilon\ \mathbb{R}^{C\times L}$, where each $x_l$ at time step $l$ is a vector of dimension $C$. The task is to forecast $T$ future values $\left\{ x_{L+1},...,x_{L+T} \right\}$. In this work, the proposed FSatten and SOatten are applied to two SOTA Transformers to compare with conventional attention, as shown in Figure \ref{MTSF Transformer} Left: (1) Variate Transformer, iTransformer \cite{liu2023itransformer}, and (2) Temporal Transformer, PatchTST \cite{nie2022time}. Many subsequent approaches \cite{zhou2023one} \cite{jin2023time} are based primarily on these two mainstream architectures. Detailed illustrations are as follows:

\subsection{Temporal Transformer}
The initial Transformer-based MTS models take time steps as tokens and apply temporal attention between them. Subsequent works demonstrated that temporal attention at a sub-series level with fewer tokens is more effective and can greatly reduce the complexity. PatchTST \cite{nie2022time} provides a general paradigm of the Temporal Transformer at the sub-series level. In PatchTST, each of the $C$ variates $X_{1:L}^{\left(i\right)}= \left\{ x_1^{\left(i\right)},\ldots x_L^{\left(i\right)} \right\} \epsilon\ \mathbb{R}^{1\times L}$ is converted to sub-series Patches $X_P^{\left(i\right)}=\left\{ x_1^{\left(i\right)},\ldots x_N^{\left(i\right)}\right\} \epsilon\ \mathbb{R}^{P\times N}$, where $N\ =\ \left\lfloor\frac{\left(L-P\right)}{S}\right\rfloor\ +2$, $P$ is length of patches and $S$ is the stride - the non-overlapping region between two consecutive patches. Temporal attention is applied to capture the dependencies between patches of each variate. The $X_P^{\left(i\right)}$ is first embedded to tokens $Z_P^{\left(i\right)}\ =\ W_H^PX_P^{\left(i\right)}$, Where $W_H^P\ \epsilon\ \mathbb{R}^{D\times P}$ and $D$ is the number of dimensions. Then the attention weight is calculated: 
\begin{equation}
 A _ { h } ^ { ( i ) } = S o f m a x ( \frac { ( ( Z _ { P } ^ { ( i ) } ) ^ { T } W _ { h } ^ { Q } ) ( ( Z _ { P } ^ { ( i ) } ) ^ { T } W _ { h } ^ { K } ) ^ { T } } { \sqrt { d _ { K } } } ) ( ( Z _ { P } ^ { ( i ) } ) ^ { T } W _ { h } ^ { V } )
\end{equation}

Where  $W _ { h } ^ { \left\{ Q , K , V \right\} }  \epsilon\ \mathbb{R}^{D\times\frac{D}{H}}$ and $H$ is number of attention heads. The mapping to latent space is $\theta_P = \left\{ W_H^P,{\ W}_H^{ \left\{ Q , K , V \right\} } \right\}$.

\subsection{Variate Transformer}
The Temporal Transformer directly follows the paradigm in NLP. But unlike natural language, MTS has multiple parallel sequence inputs. Variate Transformer explicitly models the complex correlations between variable sequences. Typically, iTrasformer \cite{liu2023itransformer} embed the whole time series of each variate $X^{(i)}$ independently into a (variate) token as $Z_V= XW_H^V$, where $W_H^V\epsilon\ \mathbb{R}^{L\times D}$. Then it adopts attention to multivariate correlations as follows:
\begin{equation}
A_h = Sofmax(\frac{({Z_VW}_h^Q){(Z_VW_h^K)}^T}{\sqrt{d_K}}) (Z_VW_h^V)
\end{equation}

The mapping to latent space of variate attention is $\theta_V\ =\ \left\{ W_H^V\ ,{\ W}_H^{\left\{ Q , K , V \right\}} \right\}$.

Whether temporal or variate as mentioned above, both transform sequences of MTS to a latent space to provide the dependency pattern between sequences. The point of our research is to demonstrate whether the mapping to latent space under conventional attention is optimal or if we can find a better one for MTSF.

\section{FSatten}

\begin{figure}[htbp]
    \vspace{-4mm}
    \centering
    \includegraphics[width=\textwidth]{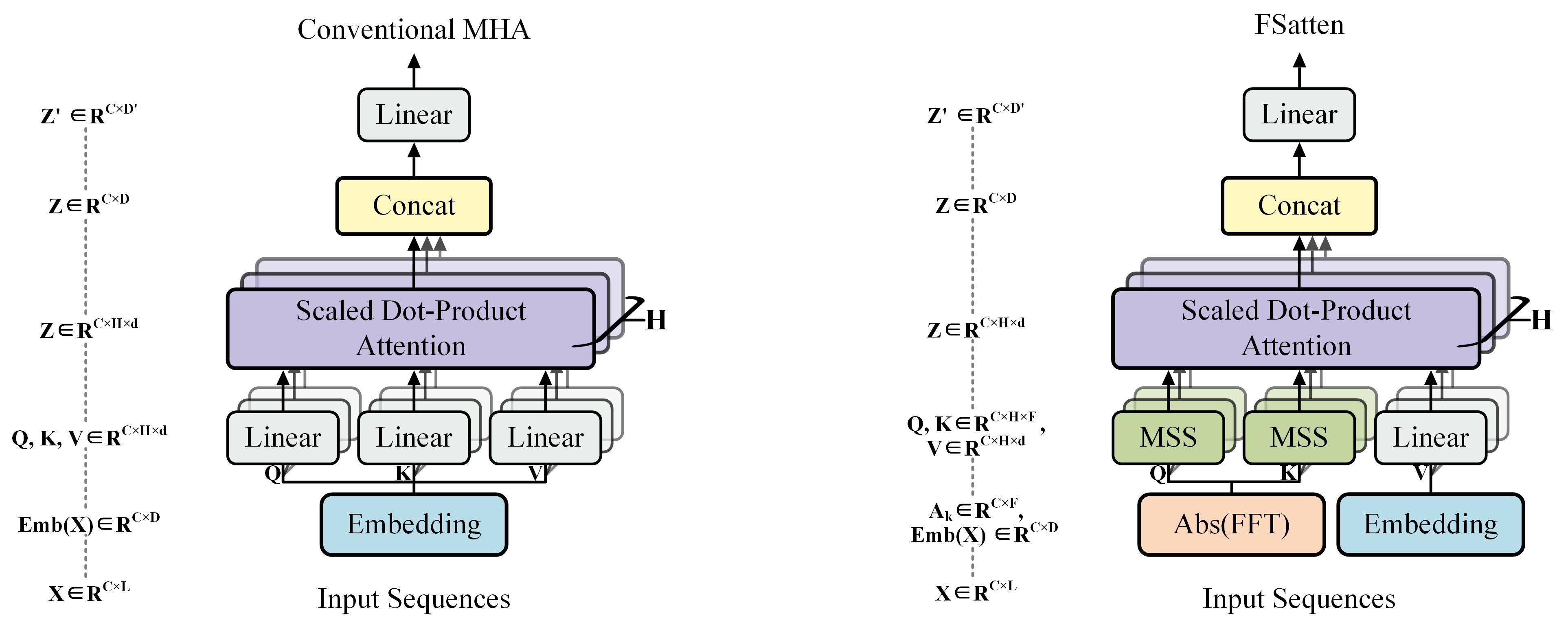}
    \caption{(left) Multi-Head Attention. (right) FSatten. On the left side of the figures is the shape of the data at each stage, and adding batch size to the front is the shape in training}
    \label{FSatten method}
    \vspace{-4mm}
\end{figure}

FSatten is an innovative attention mechanism that we propose to explore the effectiveness of conventional attention. The intuitive difference from the conventional attention, as depicted in Figure \ref{FSatten method}, is that FSatten replaces the embedding by a Fourier transform and the linear projection for the query and key by a proposed MSS.

\subsection{Workflow}
We apply the FSatten to Variate Transformer for MTSF. As shown in Figure \ref{FSatten method} right, each discrete variate sequence of the input $X$ is first transformed by the Fast Fourier Transform (FFT) \cite{brigham1967fast}, which efficiently computes the Discrete Fourier Transform (DFT) from the time domain to the complex frequency domain as:
\begin{equation}
 X _ { k } ^ { F } = \sum _ { t = 0 } ^ { L } X e ^ { - i ( 2 \pi / l ) k t } , 0 \leq k \leq F
\end{equation}
Here, $i$ is the imaginary unit, and the exponential term represents the Fourier basis associated with the different $k$ frequencies. The value of $F$ is typically half the number of data points $L$ in FFT. According to our consideration, correlations can be made up of associated frequency components with different phases. Thus, we extracted the amplitudes of different frequencies from the complex domain as follows:
\begin{equation}
 A _ { k } = | X _ { k } ^ { F } | = \sqrt { R e ( X _ { k } ^ { F } ) ^ { 2 } + I m ( X _ { k } ^ { F } ) ^ { 2 } }
\end{equation}
Where $Re$ represents the real part of $X_k^F$ and $Im$ represents the imaginary part. We then apply the MSS module for the projection of queries and keys, replacing the conventional linear projection. The aim is to compare the learnable latent space for generating attention weights with the fixed frequency domain space. Since predictions are made in the time domain, the embedding and linear projection for the value remain unchanged.
\begin{equation}
     \begin{matrix} Q = M S S _ { Q } ( A _ { k } ),\ K = M S S _ { K } ( A _ { k } ),\ V = L i n e a r _ { V } ( E m b ( X ) ) \end{matrix}
\end{equation}
After the multi-head dot product, the subsequent Feed-Forward Network (FFN) provides complicated representations by adding random noise for each variate token \cite{hornik1991approximation}.
\subsection{MSS}

\begin{figure}[htbp]
    \vspace{-4mm}
    \centering
    \includegraphics[width=0.7\textwidth]{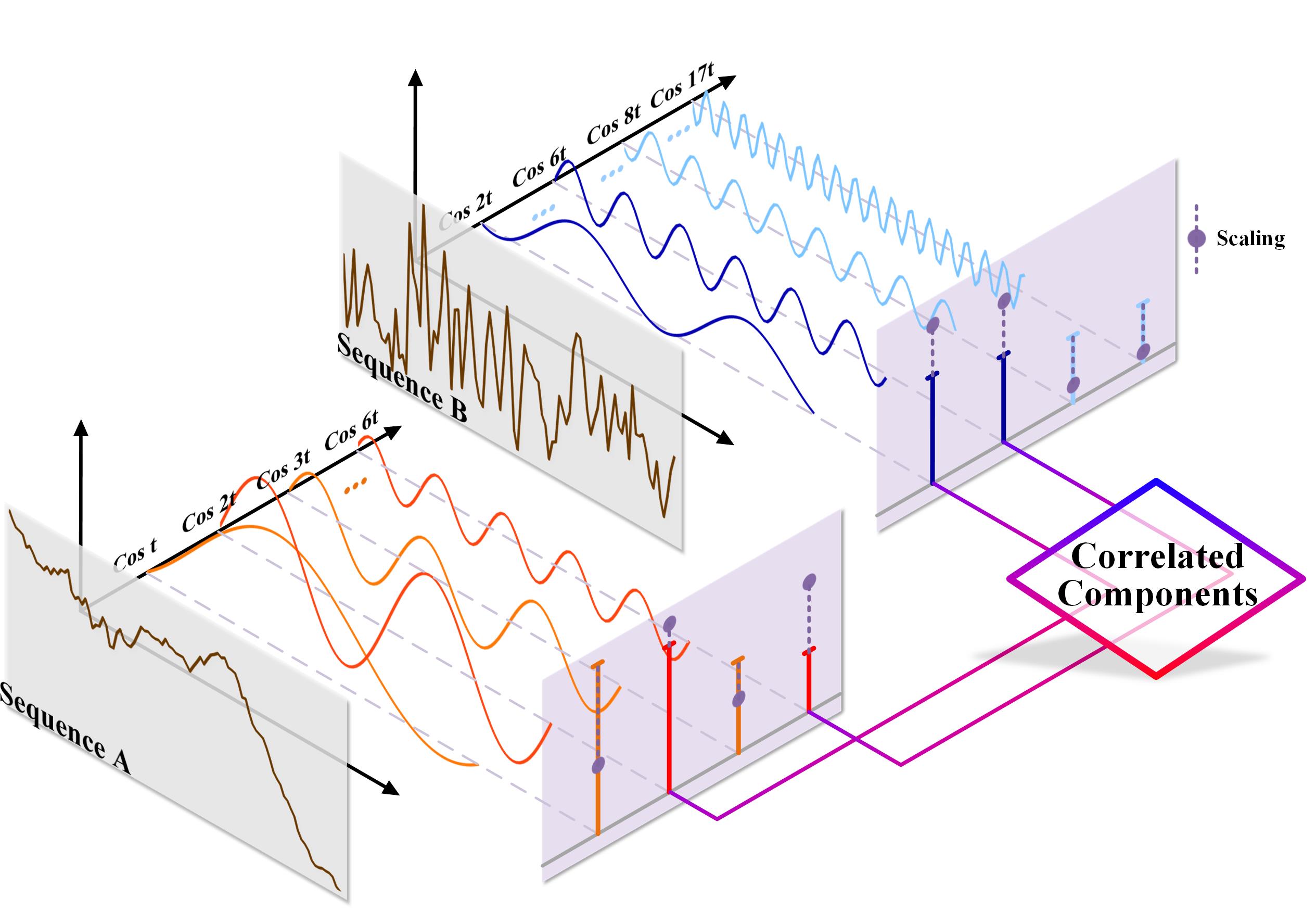}
    \caption{Multi-head Spectrum Scaling. After the Fast Fourier Transform (FFT), the correlated frequency components within the frequency domain between A and B are determined by scaled amplitude values as indicated by the \color{darkpurple}{purple} points.
}
    \label{MSS method}
    \vspace{-4mm}
\end{figure}

Although we remove the phase interference, it enables us to capture both synchronous and asynchronous associations. A more critical problem, however, is identifying the accurate associated frequency components. This is due to significant differences in amplitude values and information intensity for the same frequency across different sequences. As illustrated in Figure \ref{MSS method}, the frequency from sequence B that is potentially associated with sequence A may not represent the most significant periodic characteristics of sequence A. To efficiently obtain potentially correlated frequency components, we designed a Multi-head Spectrum Scaling (MSS) projection for queries and keys. For each attention head $h$, we scale the amplitude value $A_k$ of each frequency dimension using the Hadamard Product: 
\begin{equation}
    M S S ( A _ { k } ) = A _ { k } \circ W _ { h }
\end{equation}
where $ W_h\epsilon\ \mathbb{R}^{C\times F}$. Therefore, MSS uses $H$ different $W_h$ matrices to map the $A_k$. After learnable scaling, some frequency components can adaptively align with potentially correlated components from other sequences so that more accurate dependency patterns between sequences can be found in the subsequent dot product attention. From another consideration of maintaining the orthogonality of the frequency bias, we replace the fully connected projection, which might alter the angles between vectors and disrupt the orthogonality. Ablation experiments detailed in Section \ref{abaltions}, as well as those applied to the subsequent SOatten, demonstrate the effectiveness of MSS.

\section{SOatten}
\textbf{Motivation.} Indeed, experiments show that FSatten outperforms conventional attention, but its performance across six real-world datasets exhibits significant variance, as shown in Figure \ref{radio} and Table \ref{Average Results}. This suggests that a fixed frequency domain mapping may not be universally applicable. Determining the optimal configuration manually for each scenario is challenging, given the numerous unexplained physical transformations. Therefore, we aim to further develop FSatten to design a method with better generalization capabilities. Furthermore, FSatten may not be fully compatible with Temporal Transformers, considering that sequences derived from a single variate naturally exhibit identical periodic frequencies.

\subsection{Method}

\begin{figure}[htbp]
    \centering
    \includegraphics[width=\textwidth]{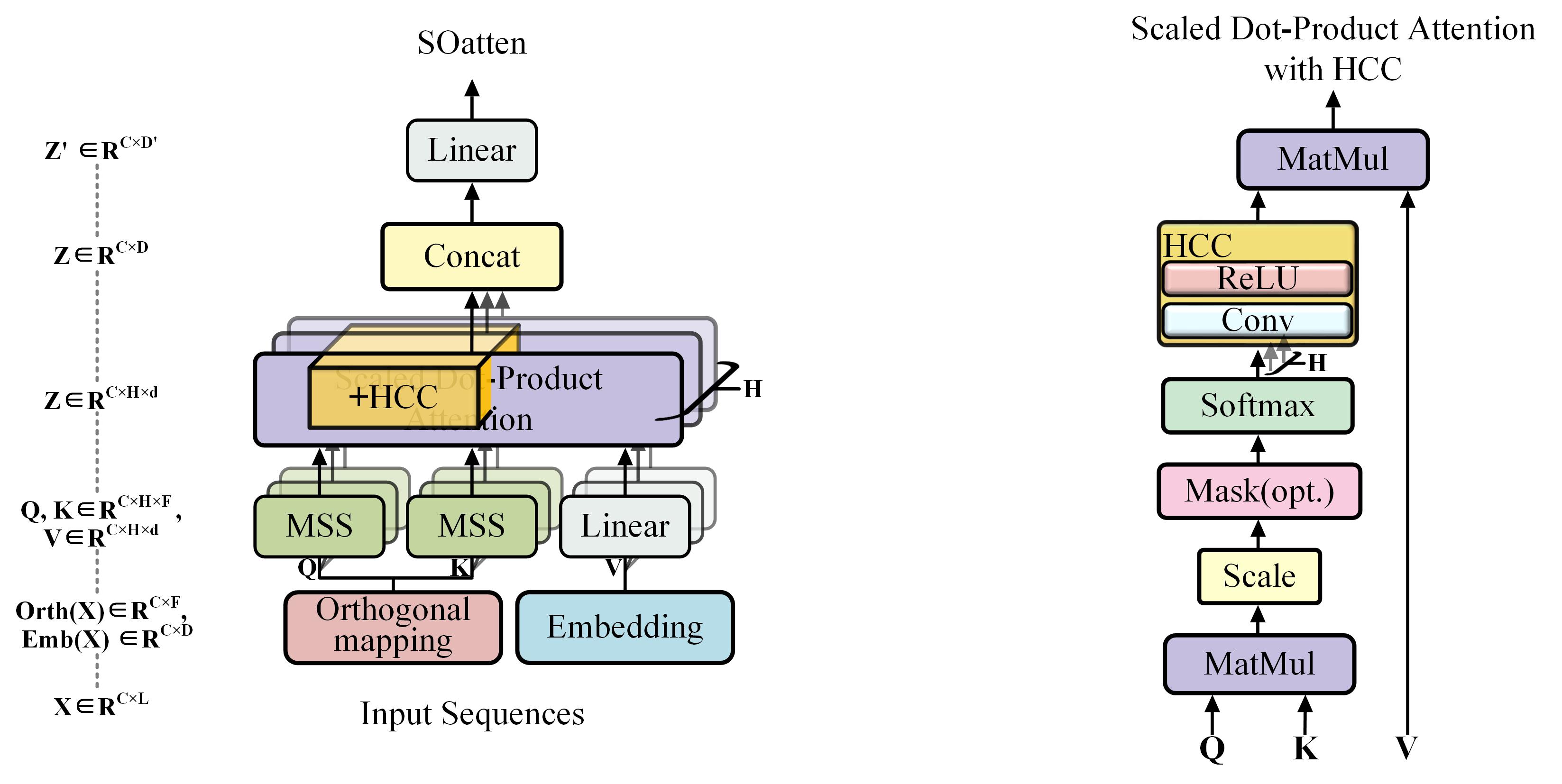}
    \caption{(left) SOatten. (right) Scaled Dot-Product Attention with HCC. On the left side of the Soatten is the shape of the data at each stage, and adding batch size to the front is the shape in training.
}
    \label{SOatten method}
\end{figure}

This is a research direction with many possible approaches, but the most straightforward idea is to extend by leveraging the orthogonality of the Fourier transform. Therefore, we propose SOatten, as depicted in Figure \ref{SOatten method}, which improves the frequency domain to a more general orthogonal domain through a learnable orthogonal transformation, described as:
\begin{equation}
O r t h ( X ) = W ^ { T } X ,\ w h e r e \ W ^ { T } W = I
\end{equation}
In fact, the orthogonality of the learnable space is not guaranteed during backpropagation for parameter updates. We could apply a measure such as QR decomposition, but this could result in the loss of some gradient information. We made a different trade-off for MTSF, considering that the scale of models that match the size of the dataset is usually not very large, with relatively few layers. Therefore, we only perform orthogonal initialization for the embedding, rather than enforcing a completely orthogonal space during backpropagation. 
Subsequently, SOatten also applies the MSS projection for query and key and applies linear projection for the value. Additionally, in dot product attention, we propose a Head Coupling Convolution (HCC) module operating on the heads of attention weights, which serves as an important 'guidance' for the mapping space learning of SOatten.

\subsection{HCC}
FSatten provides sequence dependencies based on explicit spectral information, but extending this to a learnable orthogonal space makes it difficult to effectively determine valid characteristics as a defined periodicity in FSatten. In other words, data-driven approaches that learn an effective orthogonal space without any restrictions have requirements for the size and distribution of the dataset.

I propose a general enhancement method called Head Coupling Convolution (HCC), which leverages the constraint of similarity between neighboring sequences to guide the model in exploring feature spaces. Specifically, HCC involves performing convolution operations on the attention weights within the dot product attention mechanism as:
\begin{equation}
    A t t e n = S o f t m a x ( \frac{Q K ^ { T }}{\sqrt { d _ { K } }} )
\end{equation}
\begin{equation}
    H C C ( A t t e n ) = R e L U \left\{ C o n v _ { H \rightarrow H } ( A t t e n , stride=1,kernel\_size=K ) \right\}
\end{equation}
Where ${Conv}_{H\rightarrow H}$ is channel fusion convolution that maps from $H$ heads to $H$ heads and padding is necessary for keeping the size of the weight matrix. 

For most time-series data, contrastive learning methods \cite{yue2022ts2vec} \cite{kiyasseh2021clocs} \cite{yeche2021neighborhood} \cite{tonekaboni2021unsupervised} have demonstrated the effectiveness of assumption: neighboring similarity, the similarity between sequences of the same time series decreases as the time lag increases. In fact, similar variates are arranged together in most datasets (see details in Appendix \ref{a2}). 

By applying a convolution operation to the attention weights, more critical correlated patterns between local neighboring sequences are extracted, guiding the parameter updates in the feature space during backpropagation. The diverse features extracted by $H$ heads are all predicated on neighboring similarity, multi-head coupling helps to obtain more precise associative features than single-channel convolution.

\section{Experiments}
\textbf{Navigation.} In section \ref{EXP LMTSF}, We show the forecasting results of FSatten and SOatten. In section \ref{abaltions}, I conduct ablation studies to verify the effectiveness of MSS and HCC. In section \ref{visualization}, We make a visualization analysis for FSatten and SOatten, comparing them with conventional attention. Hyperparameter sensitivity analysis is presented in Section \ref{hyperparameter}. Moreover, model efficiency analysis is presented in Appendix \ref{c3}.

c\textbf{Setting.} We extensively evaluate the proposed FSatten and SOatten on six real-world datasets, including ECL, ETT (4 subsets), Exchange, Traffic, Weather \cite{wu2021autoformer}, and Solar-Energy \cite{lai2018modeling}. Detailed dataset descriptions are provided in Appendix \ref{b1}. We choose 9 well-known forecasting models as our baselines, including (1) Transformer-based methods: Autoformer \cite{wu2021autoformer}, FEDformer \cite{zhou2022fedformer}, Crossformer \cite{zhang2022crossformer}, PatchTST \cite{nie2022time}, and iTransformer \cite{liu2023itransformer}; (2) Linear-based methods: DLinear \cite{zeng2023transformers}, TiDE \cite{das2023long}; and (3) CNN-based methods: SCINet \cite{liu2022scinet}, TimesNet \cite{wu2022timesnet}. The experimental setting is the same as in iTransformer \cite{liu2023itransformer}, that the input length $L=96$ and the output length and $T=\left\{96,192,336,720\right\}$.

\subsection{Long-term MTSF}
\label{EXP LMTSF}

\begin{table*}[htbp]
\centering
\caption{Multivariate forecasting results. Results are averaged from all prediction lengths. The input sequence length $L=96$ and the prediction lengths $T= \left\{ 96,192,336,720 \right\}$, which is consistent with the experimental setting in iTransformer \cite{liu2023itransformer}. The {\color{red}\textbf{red}} is the best and {\color{blue}\underline{blue}} is the second. Full results are listed in Appendix \ref{c1}.}
\label{Average Results}
\resizebox{\textwidth}{!}{

\begin{tabular}{c|c|cccc|cc|cc|cc|cc|cc|cc|cc|cc|cc}
\toprule
\multicolumn{2}{c|}{\multirow{2}*{Models}} 
&\multicolumn{2}{c}{\textbf{SOatten(V)}}&\multicolumn{2}{c|}{\textbf{FSatten}}&\multicolumn{2}{c|}{iTransformer}&\multicolumn{2}{c|}{PatchTST}&\multicolumn{2}{c|}{Crossformer}&\multicolumn{2}{c|}{TiDE}&\multicolumn{2}{c|}{TimesNet}&\multicolumn{2}{c|}{DLinear}&\multicolumn{2}{c|}{SCINet}&\multicolumn{2}{c|}{FEDformer}&\multicolumn{2}{c}{Autoformer}\\

\multicolumn{2}{c|}{~} &\multicolumn{2}{c}{\textbf{(Ours)}}&\multicolumn{2}{c|}{\textbf{(Ours)}}&\multicolumn{2}{c|}{(2024)}&\multicolumn{2}{c|}{(2023)}&\multicolumn{2}{c|}{(2023)}&\multicolumn{2}{c|}{(2023)}&\multicolumn{2}{c|}{(2023)}&\multicolumn{2}{c|}{(2023)}&\multicolumn{2}{c|}{(2022)}&\multicolumn{2}{c|}{(2022)}&\multicolumn{2}{c}{(2021)}\\
\cmidrule{3-24}
\multicolumn{2}{c|}{Metric}&MSE&MAE&MSE&MAE&MSE&MAE&MSE&MAE&MSE&MAE&MSE&MAE&MSE&MAE&MSE&MAE&MSE&MAE&MSE&MAE&MSE&MAE\\
\midrule

\multicolumn{2}{c|}{ETTm1}&{\color{blue}\underline{0.394}}&{\color{blue}\underline{0.402}}&0.394&0.405&0.407& 0.410& {\color{red}\textbf{0.387}}& {\color{red}\textbf{0.400}}& 0.513& 0.496& 0.419& 0.419& 0.400& 0.406& 0.403& 0.407& 0.485& 0.481& 0.448& 0.452& 0.588& 0.517 \\
\midrule

\multicolumn{2}{c|}{ETTm1}&0.287&{\color{blue}\underline{0.331}}&{\color{blue}\underline{0.286}}&0.331&0.288& 0.332&  {\color{red}\textbf{0.281}}& {\color{red}\textbf{0.326}}& 0.757& 0.610& 0.358& 0.404& 0.291& 0.333& 0.350& 0.401& 0.571& 0.537& 0.305& 0.349&  0.327& 0.371 \\
\midrule

\multicolumn{2}{c|}{ETTh1}&0.447&{\color{blue}\underline{0.440}}&{\color{blue}\underline{0.446}}&{\color{red}\textbf{0.439}}&0.454& 0.447& 0.469& 0.454& 0.529& 0.522& 0.541& 0.507& 0.458& 0.450& 0.456& 0.452& 0.747& 0.647& {\color{red}\textbf{0.440}}& 0.460&  0.496& 0.487 \\
\midrule

\multicolumn{2}{c|}{ETTh2}&{\color{red}\textbf{0.379}}&{\color{red}\textbf{0.405}}&{\color{blue}\underline{0.381}}&{\color{blue}\underline{0.407}}& 0.383& 0.407&  0.387& 0.407& 0.942& 0.684& 0.611& 0.550& 0.414& 0.427& 0.559& 0.515& 0.954& 0.723& 0.437& 0.449&  0.450& 0.459 \\
\midrule

\multicolumn{2}{c|}{ECL}&{\color{blue}\underline{0.166}}&{\color{blue}\underline{0.259}}&{\color{red}\textbf{0.162}}&{\color{red}\textbf{0.257}}&0.178& 0.270&  0.216& 0.304& 0.244& 0.334& 0.251& 0.344& 0.192& 0.295& 0.212& 0.300& 0.268& 0.365& 0.214& 0.327&  0.227& 0.338 \\
\midrule

\multicolumn{2}{c|}{Exchange}&{\color{blue}\underline{0.359}}&{\color{blue}\underline{0.404}}&0.363&0.406&0.360& {\color{red}\textbf{0.403}}&  0.367& 0.404& 0.940& 0.707& 0.370& 0.413& 0.416& 0.443& {\color{red}\textbf{0.354}}& 0.414& 0.750& 0.626& 0.519& 0.429&  0.613& 0.539 \\
\midrule

\multicolumn{2}{c|}{Traffic}&{\color{blue}\underline{0.437}}&{\color{blue}\underline{0.286}}&0.477&0.291&{\color{red}\textbf{0.428}}& {\color{red}\textbf{0.282}}&  0.555& 0.362& 0.550& 0.304& 0.760& 0.473& 0.620& 0.336& 0.625& 0.383& 0.804& 0.509& 0.610& 0.376&  0.628& 0.379 \\
\midrule

\multicolumn{2}{c|}{Weather}&{\color{red}\textbf{0.245}}&{\color{red}\textbf{0.273}}&{\color{blue}\underline{0.249}}&{\color{blue}\underline{0.275}}&0.258& 0.279&  0.259& 0.281& 0.259& 0.315& 0.271& 0.320& 0.259& 0.287& 0.265& 0.317& 0.292& 0.363& 0.309& 0.360&  0.338& 0.382 \\
\midrule

\multicolumn{2}{c|}{Solar-Energy}&{\color{red}\textbf{0.229}}&{\color{blue}\underline{0.261}}&{\color{blue}\underline{0.230}}&{\color{red}\textbf{0.259}}&0.233& 0.262&  0.270& 0.307& 0.641& 0.639& 0.347& 0.417& 0.301& 0.319& 0.330& 0.401& 0.282& 0.375& 0.291& 0.381&  0.885& 0.711 \\

\bottomrule
\end{tabular}
}
\end{table*}

Compared to the baselines presented in Table  \ref{Average Results}, FSatten, based on the Variate Transformer, shows overall better forecasting performance than the SOTA, which uses conventional attention mechanisms. Particularly for datasets with more pronounced periodicity, such as on ECL, FSatten significantly improves performance by an overall 9.0\% compared to SOTA and exhibits greater stability for longer prediction sequences. These improvements demonstrate that FSatten effectively captures the accurate correlation at the same frequency, which is more suitable for application in Transformers for MTSF.

\begin{table*}[htbp]
\centering
\caption{Forecasting results of SOatten under Temporal and Variate Transformers. \textbf{Bolded} results are superior to conventional attention. Results are averaged from all prediction lengths. Full results are listed in Appendix \ref{c1}.}
\label{SOatten Average Result}
\begin{tabular}{c|c|cccc|cccc}
\toprule
\multicolumn{2}{c|}{\multirow{2}*{Models}} 
&\multicolumn{4}{c|}{Temporal Transformer}&\multicolumn{4}{c}{Variate Transformer}\\

\multicolumn{2}{c|}{~} &\multicolumn{2}{c}{\textbf{SOatten(T)}}&\multicolumn{2}{c|}{PatchTST}&\multicolumn{2}{c}{\textbf{SOatten(V)}}&\multicolumn{2}{c}{iTransformer}\\
\cmidrule{3-10}
\multicolumn{2}{c|}{Metric}&MSE&MAE&MSE&MAE&MSE&MAE&MSE&MAE\\
\midrule

\multicolumn{2}{c|}{ETTm1}&\textbf{0.380}&\textbf{0.395}&0.387& 0.400&\textbf{0.394}&\textbf{0.402}&0.407& 0.410\\
\multicolumn{2}{c|}{ETTm2}&\textbf{0.280}&0.326&0.281& 0.326&\textbf{0.287}&\textbf{0.331}&0.288& 0.332\\
\multicolumn{2}{c|}{ETTh1}&\textbf{0.451}&\textbf{0.441}&0.469& 0.454&\textbf{0.447}&\textbf{0.440}&0.454& 0.447\\
\multicolumn{2}{c|}{ETTh2}&\textbf{0.366}&\textbf{0.395}&0.387& 0.407&\textbf{0.379}&\textbf{0.405}&0.383& 0.407\\
\multicolumn{2}{c|}{ECL}&\textbf{0.199}&\textbf{0.282}&0.216 &0.304&\textbf{0.166}&\textbf{0.259}&  0.178 &0.270\\
\multicolumn{2}{c|}{Exchange}&\textbf{0.360}&\textbf{0.398}&0.367 &0.404&\textbf{0.359}& 0.404& 0.360 &0.403\\
\multicolumn{2}{c|}{Traffic}&\textbf{0.492}&\textbf{0.310}&0.555& 0.362&0.437& 0.286&  0.428 &0.282\\
\multicolumn{2}{c|}{Weather}&\textbf{0.256}&\textbf{0.280}&0.259& 0.281&\textbf{0.245}&\textbf{0.273}&  0.258 &0.279\\
\multicolumn{2}{c|}{Solar-Energy}&\textbf{0.259}&\textbf{0.284}&0.270 &0.307&\textbf{0.229}&\textbf{0.261}&  0.233& 0.262\\
\bottomrule
\end{tabular}
\end{table*}

Periodicity is one of the most fundamental characteristics of time series, but not all datasets exhibit strong periodicity. Thus, as a more general approach that can be adapted to both Temporal and Variate transformers, SOatten achieves more comprehensive improvements relative to FSatten across different scenarios. We can observe in Table \ref{SOatten Average Result} that, although each Transformer excels on certain datasets, SOatten consistently outperforms conventional attention mechanisms, regardless of the architecture. Of course, FSatten can provide superior performance for datasets that are known to exhibit stronger periodicity. 

\subsection{Ablation Studies}
\label{abaltions}

\begin{figure}[htbp]
    \vspace{-4mm}
    \centering
    \includegraphics[width=\textwidth]{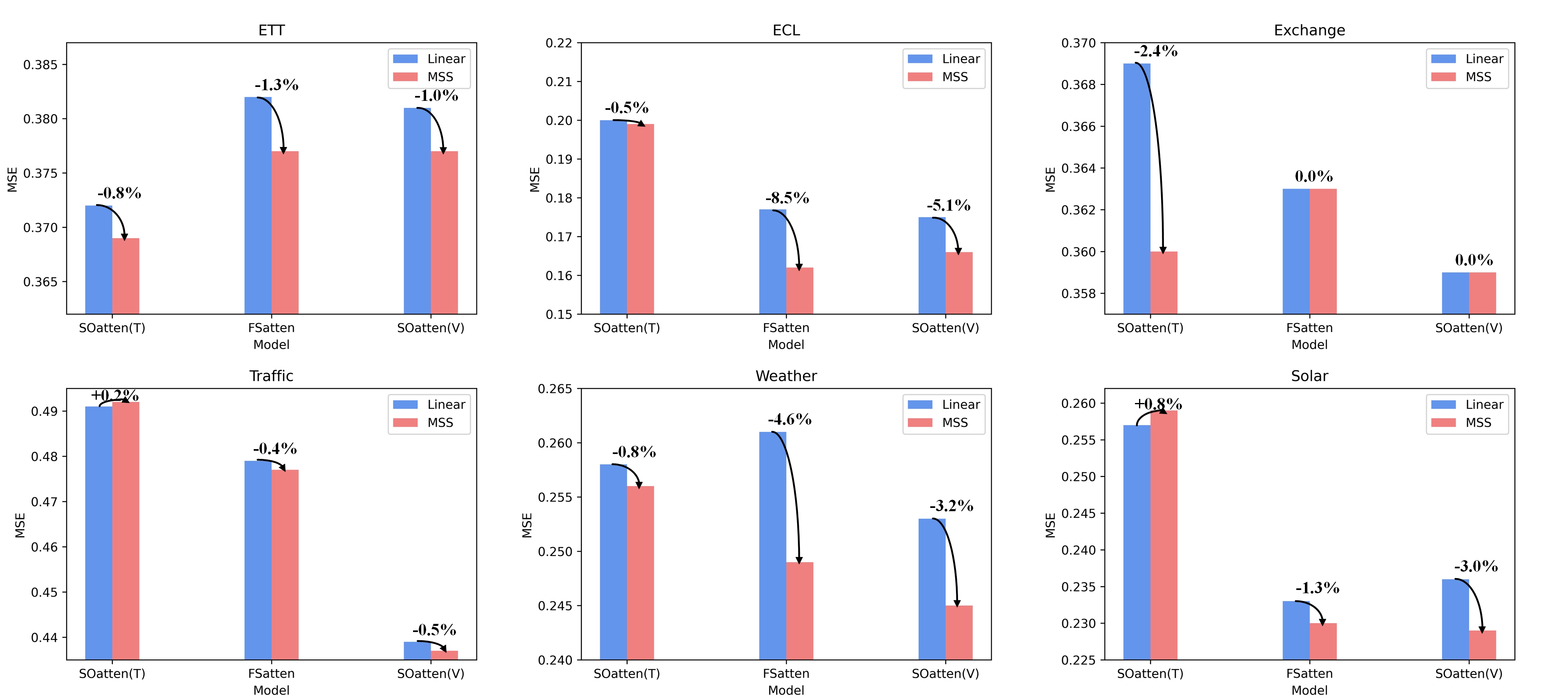}
    \caption{Ablation for the Multi-head Spectrum Scaling module. Linear represents that FSatten and SOatten use Linear projections for Query and Key. MSE scores are averaged from all prediction lengths. 
}
    \label{Ablation MSS}
\end{figure}

To validate the effectiveness of the components of FSatten and SOatten, detailed ablation experiments are conducted. The effectiveness of the MSS mapping module, used in both methods, is compared with that of applying a linear mapping to FSatten and SOatten, as shown in Figure \ref{Ablation MSS}. MSS significantly outperforms the fully connected layer, corroborating its ability to identify more accurate associated components. In Section \ref{visualization}, a visual interpretation is provided to explain in detail.

\begin{figure}[htbp]
    \vspace{-3mm}
    \centering
    \includegraphics[width=\textwidth]{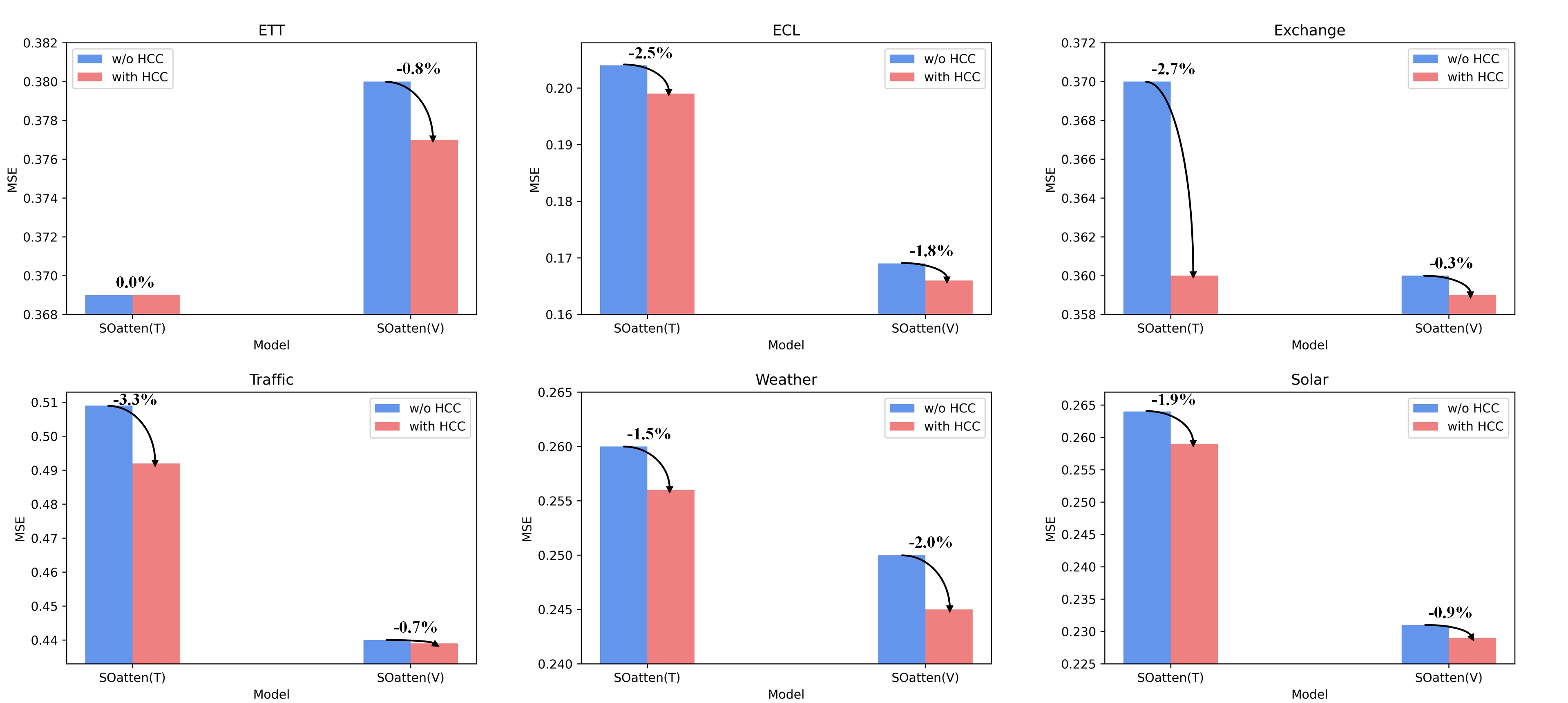}
    \caption{Ablation for the Head Coupling Convolution module. w/o represents SOatten without HCC module. MSE scores are averaged from all prediction lengths. 
}
    \label{Ablation HCC}
    \vspace{-4mm}
\end{figure}

Secondly, we validate the effectiveness of the HCC module in SOatten in Figure \ref{Ablation HCC}. The HCC is an important design for SOatten, significantly enhancing forecasting performance. Especially under the Temporal Transformer, the HCC demonstrates better generalizability. These results prove that neighboring similarity is crucial for the formation of an effective orthogonal mapping space and the generation of accurate attention weights. In Section \ref{hyperparameter}, we conduct further hyperparameter analysis on the kernel size $K$ of HCC.

\subsection{Visualized Analysis}
\label{visualization}
First, we make visualizations of generated attention matrices and analyze the advantages of the proposed two attention mechanisms. In the upper part of Figure \ref{Atten weather}, under the Variate Transformer, SOatten and FSatten generate smaller ranges but more refined weight values than the conventional attention applied by the SOTA, iTransformer (compare the value range on the right side of heatmaps). There are three main points of analysis:

\begin{figure}[htbp]
    \centering
    \includegraphics[width=\textwidth]{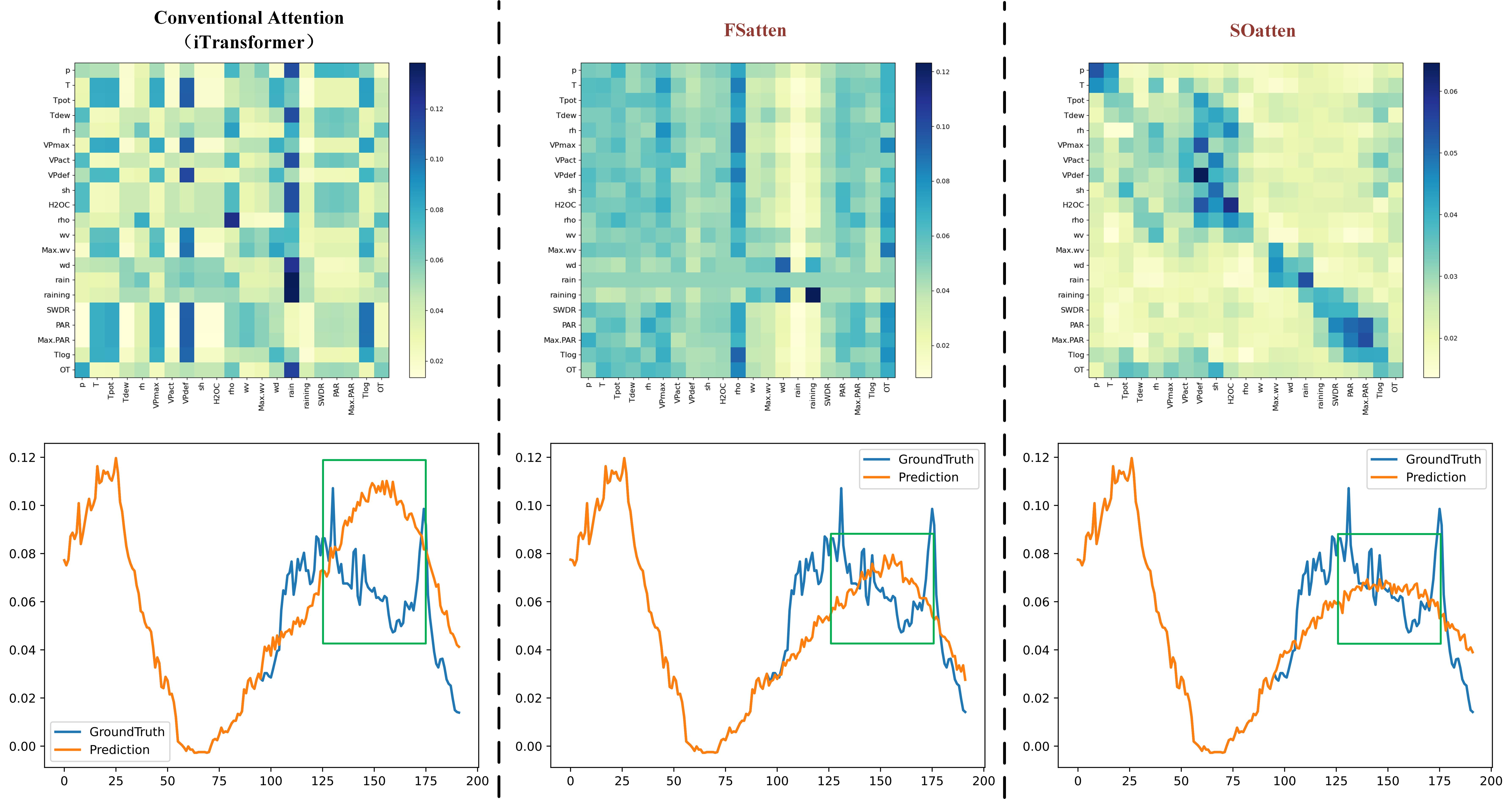}
    \caption{Attention maps and the forecasting of a few time series from Weather dataset under Variate Transformer. The attention map is calculated by averaging the attention matrices over all the heads and across all the layers.
}
    \label{Atten weather}
\end{figure}

(1) In the generated attention weight maps, the patterns of the conventional attention and FSatten show similarities, presenting dependencies that are based on the sequence periodicity. However, FSatten significantly reflects complex associations from more variable sequences, which is a benefit from the designed spectrum correlating in frequency domain space.

(2) The weights map generated by SOatten is significantly different, seemingly finding accurate dependencies based on other associated physical quantities in addition to periodicity (as seen in the upper left part of SOatten’s attention map). Furthermore, if the HCC module is not used (shown in the Appendix \ref{d}), SOatten finds new physical quantities but fails to produce a comprehensive dependency pattern, proving the effectiveness of the neighboring similarity design. 

(3) Numerical analysis of the weight matrices (in Appendix \ref{d}) shows that the proposed FSatten and SOatten are both full rank (21), while the conventional attention weight matrix is not full rank (19). The condition numbers of the weight matrices generated by FSatten (1,519) and SOatten (1,480) are much smaller than that of the conventional attention (78,596,560). These indicate that the orthogonal spaces of FSatten and SOatten are more informative and have better stability against noise than the latent space of conventional attention.

Secondly, in the lower part of Figure \ref{Atten weather}, the predictions indicate that the conventional attention mechanism's fits are poor, which appears to have learned inaccurate periodic patterns. By leveraging the frequency domain and the MSS module, FSatten finds a more accurate pattern, that combines periodic dependencies based on frequency spectrum from other variates. SOatten finds an even better pattern by combining periodicity and other key physical quantities thereby avoiding prediction errors caused by an exclusive reliance on periodicity, as shown in the green box in Figure \ref{Atten weather}. More predictions are shown in the Appendix \ref{d}.

\subsection{Hyperparameter Sensitivity}
\label{hyperparameter}
\begin{figure}[htbp]
    \centering
    \includegraphics[width=\textwidth]{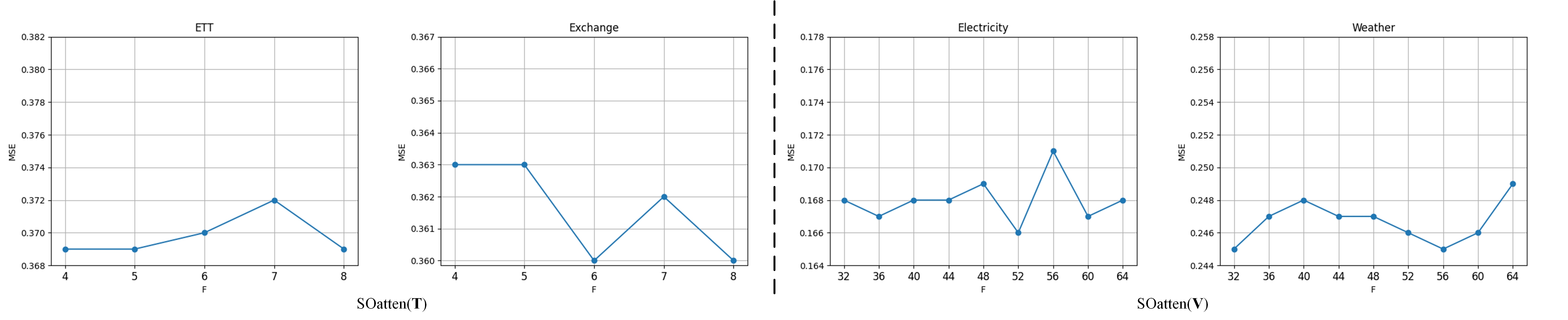}
    \caption{MSE scores averaged from all prediction lengths with varying parameter $F$ in SOatten. The left two show the results of SOatten under the Temporal Transformer, while the right two figures show the results of SOatten under the Variate Transformer. The upper limit of the coordinate values represents the performance of SOTA.
}
    \label{hyper F}
\end{figure}

Compared to conventional attention, the main hyperparameter differences are in the dimension size of the orthogonal mapping space $F$ for MSS and the kernel size $K$ for HCC, as shown in Figures \ref{FSatten method} and \ref{SOatten method}. For FSatten, the $F$ frequency components are orthogonal and are typically set to $F = (\frac{L}{2} + 1)$. In contrast, SOatten orthogonally initializes the mapping weights, and the dimension size $F$ of the mapping spaces for Query and Key are set separately from that of the Value (same with conventional attention). Sensitivity experiments on $F$, depicted in Figure \ref{hyper F}, demonstrate that the performance of the proposed attention mechanisms is not coincidental. When optimized, SOatten's $F$ is significantly smaller than the $d_Q$ and $d_K$ of conventional attention, thereby enhancing the model's efficiency to some extent (detailed efficiency analysis is presented in Appendix \ref{c3}).

\begin{figure}[htbp]
    \centering
    \includegraphics[width=\textwidth]{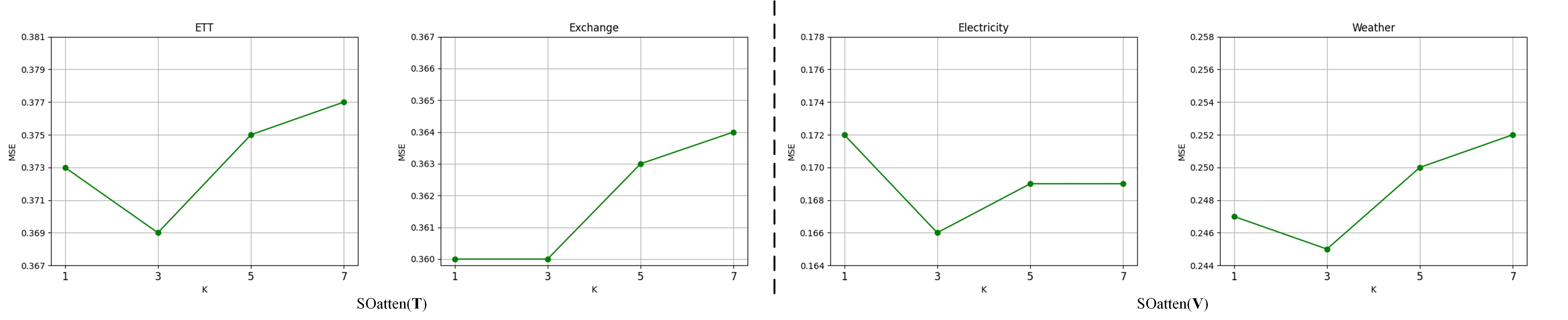}
    \caption{MSE scores averaged from all prediction lengths with varying parameter $K$ in the HCC module. The left two show the results of SOatten under the Temporal Transformer, while the right two figures show the results of SOatten under the Variate Transformer. The upper limit of the coordinate values represents the performance of SOTA.
}
    \label{hyper K}
\end{figure}

Secondly, we compared different HCC kernel sizes $K$, as shown in Figure \ref{hyper K}. The upper limit of the coordinate values represents the performance of SOTA. Under both the Temporal and Variate Transformers, the model performance using HCC with different kernel sizes is consistently better than SOTA, demonstrating the effectiveness of local neighboring similarity. However, a $3\times3$ convolutional kernel is found to be the most appropriate choice; it does not increase the complexity due to its small size.

\section{Disscussion}
For the MTSF problem, this paper proposes two innovative attention that are superior to conventional attention. We started from the frequency domain and made preliminary explorations based on the mainstream Temporal and Variate Transformers. We will soon supplement comprehensive experimental results of applying FSatten and SOatten to new architectures released this year, such as TimeXer \cite{wang2024timexer}, and the Cross Spatio-Temporal architecture (another work of ours). Constrained by our resources, we encourage the community to explore solutions to these problems together. 

we outline three promising directions for future work:

\textbf{Mapping spaces based on other physical properties}, as there exist trend, seasonality, and other characteristics in addition to periodicity.

\textbf{Combine with state-space models}, especially for datasets with a large number of variates, The advantage of current state-space models such as Mamba \cite{gu2023mamba} \cite{ahamed2024timemachine} \cite{patro2024simba} and TTT \cite{sun2024learning} is that they can compress the larger variable background into a fixed hidden state, selectively retaining the most important information under global variates, which is very important for scenarios with multiple classes of variable information, like in Traffic. Of course, this is an improved method that we will launch soon.

\textbf{Apply to other fields}, Such as CV (Computer Vision), NLP (Natural Language Processing), and motion planning, which are all directions that can explore the proposed FSatten and SOatten.


\appendix
\begin{appendix}

\section{Related Work}
\label{a}
\subsection{Transformer-based MTSF methods}
\label{a1}
As the Transformer continues to make breakthroughs in the fields of NLP and CV, applying the attention mechanism to time series analysis outperforms TCN- and RNN-based methods. LogTrans \cite{li2019enhancing} proposes LogSparse attention, which focuses on the previous step at exponential intervals in each step to break the memory bottleneck. Reformer \cite{kitaev2020reformer} employs position-sensitive hashing (LSH) to divide tokens into several buckets and perform attention within each bucket. Informer \cite{zhou2021informer} introduces ProbSparse attention, a mechanism that calculates the top-u leading queries based solely on the measured query sparsity. Autoformer \cite{wu2021autoformer} constructs a series-level connection based on the process similarity derived by series periodicity and introduces a decomposition forecasting architecture. FEDformer \cite{zhou2022fedformer} starts from the frequency domain. It selects a random subset of frequency components and multiplies them by learnable complex parameters to learn a sparse representation of time series. PatchTST \cite{nie2022time} adopts a general patch-based series representation inspired by the Vision Transformer \cite{dosovitskiy2020image}. Crossformer \cite{zhang2022crossformer} explicitly captures the cross-time and cross-variate dependencies through two-stage attention and a renovated architecture. iTransformer\cite{liu2023itransformer} embeds the time points of individual series into variate tokens which are utilized by the attention mechanism to capture multivariate correlations.
These methods have continuously improved from the model architecture and the application of Attention. However, none of them have delved into the internal workings of the attention mechanism to explore whether conventional attention is optimal for MTSF.

\subsection{Neighboring Similarity}
\label{a2}

\begin{figure}[htbp]
    \centering
    \includegraphics[width=\textwidth]{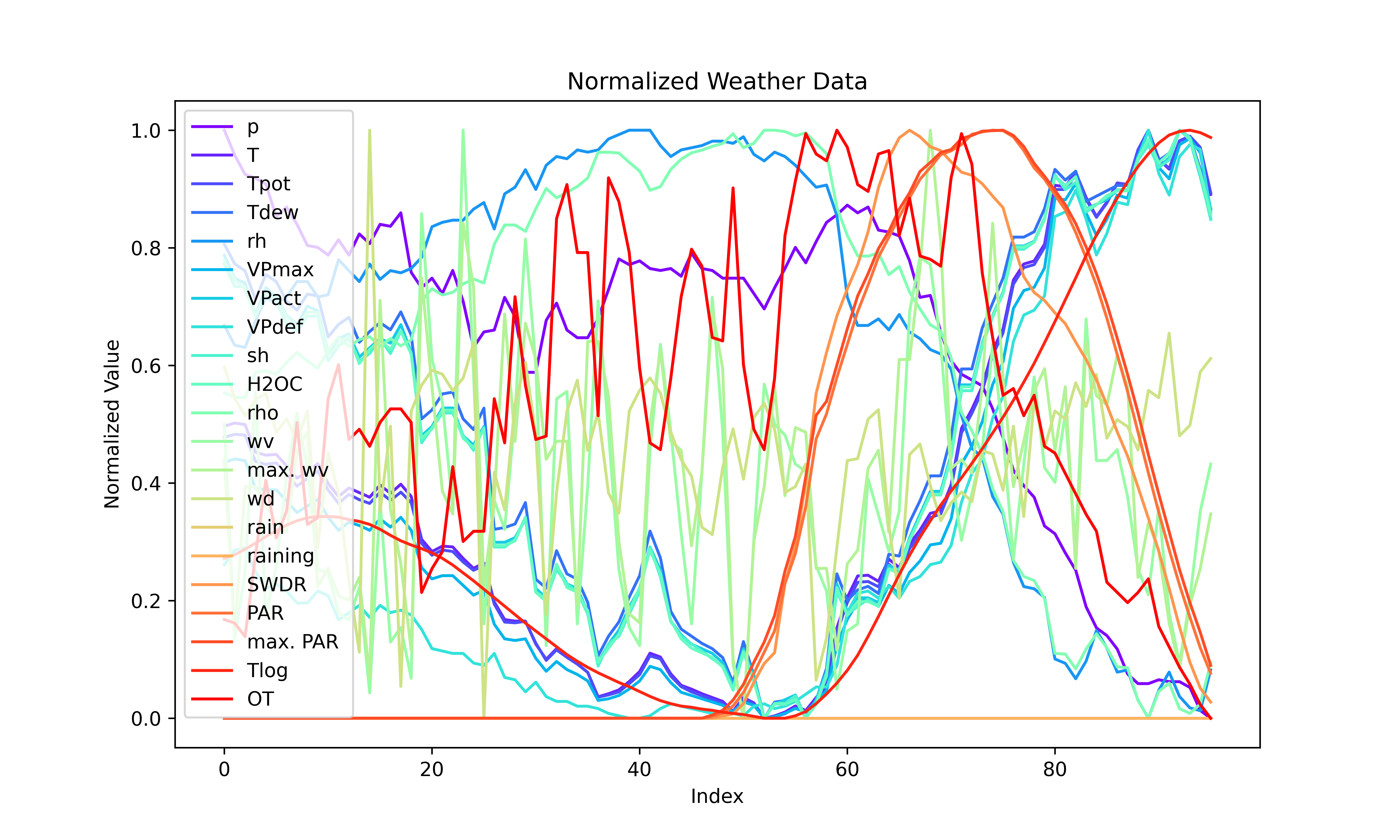}
    \caption{A partial of the weather dataset (21 variates). 
}
    \label{Weather dataset}
\end{figure}

For the sub-sequence patches segmented from the same variate, in most cases, they meet the assumption of neighboring similarity: the similarity between sequences of the same time series decreases as the time lag increases. We found that most MTS datasets such as widely used ETT, ECL, and Weather, also exhibit certain degrees of neighboring similarity between variates. As shown in Figure \ref{Weather dataset}, In the Weather dataset, adjacent variates have similar curve shapes. The attention weights and forecasting performance have verified the significance of the neighboring similarity bias for SOatten.

\section{Experimental Details}
\label{b}
\subsection{Dataset Description}
\label{b1}
We conducted experiments on six real-world datasets to evaluate the performance of the proposed FSatten and SOatten. The ETT series includes data on seven oil and load features of electricity transformers. Traffic contains hourly road occupancy rates recorded by San Francisco freeway sensors from 2015 to 2016. Weather contains meteorological observations including temperature, humidity, wind speed, and precipitation. Exchange contains panel data on daily exchange rates from 8 countries, spanning from 1990 to 2016. ECL contains hourly electricity consumption data (in kWh) for 321 clients from 2012 to 2014. Solar-Energy contains hourly solar power output data collected from 137 PV plants in Alabama in 2007. We provide the dataset details in Table \ref{datasets}.

\begin{table*}[htbp]
\centering
\captionsetup{type=table,labelsep=period,labelfont=bf} 
\caption{Detailed dataset descriptions.}
\label{datasets}
        \resizebox{\textwidth}{!}{
        \begin{tabular}{c|ccccccccc}
            \hline
            Datasets&ETTm1&ETTm2&ETTh1&ETTh2&ECL&Exchange&Traffic&Weather&Solar-Energy\\
            \hline
            Timestamps&69,680&69,680&17,420&17,420&26,340&7,588&17,544&52,696&52,560\\
            Variates&7&7&7&7&321&8&862&21&137\\
            Frequency&15min&15min&Hourly&Hourly&Hourly&Daily&Hourly&10min&10min\\
            \hline
    \end{tabular}}
\end{table*}

\subsection{Implementation Detials}
\label{b2}

All the experiments are implemented in PyTorch \cite{paszke2019pytorch} and conducted on a single NVIDIA RTX 3090 24GB GPU. To validate our approach, we only replace the conventional attention with FSatten and SOatten without changing any original parameter settings in iTransformer and PatchTST. 
PatchTST includes $3$ encoder layers with a head number of $H = 16$ and a latent space dimension of $D = 128$ and a prediction head dimension $F = 256$. For ETT, parameters of a reduced size are used ($H = 4$, $D = 16$, and  $F = 128$) to reduce the risk of overfitting. A dropout rate of $0.2$ is applied in the encoders for all experiments. iTransformer contains $\left\{2, 3, 4\right\}$ layers with a head number of $H = 8$, The dimension of series representations $D = \left\{256, 512\right\}$.The only difference is the addition of two hyperparameters $F = \left\{32,49\right\}$ and kernel size $K = \left\{1,3\right\}$ in HCC.

\section{Full Results}
\label{c}
\subsection{Long-term Forecasting}
\label{c1}
We give the full forecasting results in Table \ref{Full MTSF Result}, we compare SOatten and FSatten based on Variate Transformer with benchmarks, and we can see that SOatten and FSatten are in the top two on all six real-world datasets. Table  \ref{Full SOatten Result} shows the full results of directly comparing SOatten with the conventional attention based on both Variate Transformer and Temporal Transformer, SOatten achieved a comprehensive surpass.

\begin{table*}
\centering
\caption{Full forecasting results of MTSF. The fixed input sequence length $L=96$ and the prediction lengths $T= \left\{ 96,192,336,720 \right\}$, which is consistent with the experimental setting in iTransformer \cite{liu2023itransformer}. The {\color{red}\textbf{red}} is the best and {\color{blue}\underline{blue}} is the second. Avg means the average results of all prediction lengths.}
\label{Full MTSF Result}
\resizebox{\textwidth}{!}{
\begin{tabular}{c|c|cccc|cc|cc|cc|cc|cc|cc|cc|cc|cc}
\toprule
\multicolumn{2}{c|}{\multirow{2}*{Models}} 
&\multicolumn{2}{c}{\textbf{SOatten(V)}}&\multicolumn{2}{c|}{\textbf{FSatten}}&\multicolumn{2}{c|}{iTransformer}&\multicolumn{2}{c|}{PatchTST}&\multicolumn{2}{c|}{Crossformer}&\multicolumn{2}{c|}{TiDE}&\multicolumn{2}{c|}{TimesNet}&\multicolumn{2}{c|}{DLinear}&\multicolumn{2}{c|}{SCINet}&\multicolumn{2}{c|}{FEDformer}&\multicolumn{2}{c}{Autoformer}\\

\multicolumn{2}{c|}{~} &\multicolumn{2}{c}{\textbf{(Ours)}}&\multicolumn{2}{c|}{\textbf{(Ours)}}&\multicolumn{2}{c|}{(2024)}&\multicolumn{2}{c|}{(2023)}&\multicolumn{2}{c|}{(2023)}&\multicolumn{2}{c|}{(2023)}&\multicolumn{2}{c|}{(2023)}&\multicolumn{2}{c|}{(2023)}&\multicolumn{2}{c|}{(2022)}&\multicolumn{2}{c|}{(2022)}&\multicolumn{2}{c}{(2021)}\\
\cmidrule{3-24}
\multicolumn{2}{c|}{Metric}&MSE&MAE&MSE&MAE&MSE&MAE&MSE&MAE&MSE&MAE&MSE&MAE&MSE&MAE&MSE&MAE&MSE&MAE&MSE&MAE&MSE&MAE\\
\midrule

\multirow{5}*{\rotatebox{90}{ETTm1}}&96&{\color{red}\textbf{0.329}}&{\color{red}\textbf{0.365}}&0.331&0.369&0.334& 0.368&  {\color{blue}\underline{0.329}}& {\color{blue}\underline{0.367}}& 0.404& 0.426& 0.364& 0.387& 0.338& 0.375& 0.345& 0.372& 0.418& 0.438& 0.379& 0.419&  0.505& 0.475 \\

~&192&{\color{blue}\underline{0.370}}&{\color{blue}\underline{0.387}}&0.373&0.391&0.377& 0.391&  {\color{red}\textbf{0.367}}& {\color{red}\textbf{0.385}}& 0.450& 0.451& 0.398& 0.404& 0.374& 0.387& 0.380& 0.389& 0.439& 0.450& 0.426& 0.441&  0.553& 0.496 \\

~&336&{\color{blue}\underline{0.401}}&{\color{red}\textbf{0.407}}&0.404&0.411&0.426& 0.420& {\color{red}\textbf{0.399}}& {\color{blue}\underline{0.410}}& 0.532& 0.515& 0.428& 0.425& 0.410& 0.411& 0.413& 0.413& 0.490& 0.485& 0.445& 0.459&  0.621& 0.537 \\

~&720&0.474& {\color{blue}\underline{0.447}}& {\color{blue}\underline{0.469}}&0.447&0.491& 0.459&  {\color{red}\textbf{0.454}}& {\color{red}\textbf{0.439}}& 0.666& 0.589& 0.487& 0.461& 0.478& 0.450& 0.474& 0.453& 0.595& 0.550& 0.543& 0.490&  0.671& 0.561 \\
\cmidrule{2-24}
~&Avg&{\color{blue}\underline{0.394}}&{\color{blue}\underline{0.402}}&0.394&0.405&0.407& 0.410& {\color{red}\textbf{0.387}}& {\color{red}\textbf{0.400}}& 0.513& 0.496& 0.419& 0.419& 0.400& 0.406& 0.403& 0.407& 0.485& 0.481& 0.448& 0.452& 0.588& 0.517 \\
\midrule
\multirow{5}*{\rotatebox{90}{ETTm2}}&96&{\color{blue}\underline{0.180}}&{\color{blue}\underline{0.264}}&0.180&0.265&0.180& 0.264&  {\color{red}\textbf{0.175}}& {\color{red}\textbf{0.259}}& 0.287& 0.366& 0.207& 0.305& 0.187& 0.267& 0.193& 0.292& 0.286& 0.377& 0.203& 0.287&  0.255& 0.339 \\

~&192&{\color{blue}\underline{0.245}}&{\color{blue}\underline{0.306}}&0.245&0.307&0.250& 0.309&  {\color{red}\textbf{0.241}}& {\color{red}\textbf{0.302}}& 0.414& 0.492& 0.290& 0.364& 0.249& 0.309& 0.284& 0.362& 0.399& 0.445& 0.269& 0.328&  0.281& 0.340 \\

~&336&0.312&0.349&{\color{blue}\underline{0.308}}&{\color{blue}\underline{0.347}}&0.311& 0.348&  {\color{red}\textbf{0.305}}& {\color{red}\textbf{0.343}}& 0.597& 0.542& 0.377& 0.422& 0.321& 0.351& 0.369& 0.427& 0.637& 0.591& 0.325& 0.366&  0.339& 0.372 \\

~&720&{\color{blue}\underline{0.411}}&0.406& 0.412&{\color{blue}\underline{0.405}}&0.412& 0.407&  {\color{red}\textbf{0.402}}&  {\color{red}\textbf{0.400}}& 1.730& 1.042& 0.558& 0.524& 0.408& 0.403& 0.554& 0.522& 0.960& 0.735& 0.421& 0.415&  0.433& 0.432 \\
\cmidrule{2-24}
~&Avg&0.287&{\color{blue}\underline{0.331}}&{\color{blue}\underline{0.286}}&0.331&0.288& 0.332&  {\color{red}\textbf{0.281}}& {\color{red}\textbf{0.326}}& 0.757& 0.610& 0.358& 0.404& 0.291& 0.333& 0.350& 0.401& 0.571& 0.537& 0.305& 0.349&  0.327& 0.371 \\
\midrule
\multirow{5}*{\rotatebox{90}{ETTh1}}&96&{\color{blue}\underline{0.383}}&{\color{red}\textbf{0.400}}& 0.384&	{\color{blue}\underline{0.401}}&0.386& 0.405&  0.414& 0.419& 0.423& 0.448& 0.479& 0.464& 0.384& 0.402& 0.386&  0.400& 0.654& 0.599& {\color{red}\textbf{0.376}}& 0.419&  0.449& 0.459 \\

~&192&{\color{blue}\underline{0.440}}&0.433&0.434&{\color{red}\textbf{0.429}}&0.441& 0.436& 0.460& 0.445& 0.471& 0.474& 0.525& 0.492& 0.436&  0.429& 0.437& {\color{blue}\underline{0.432}}& 0.719& 0.631& {\color{red}\textbf{0.420}}& 0.448&  0.500& 0.482 \\

~&336&{\color{blue}\underline{0.475}}&{\color{red}\textbf{0.449}}&0.481&{\color{blue}\underline{0.453}}&0.487& 0.458&  0.501& 0.466& 0.570& 0.546& 0.565& 0.515& 0.491& 0.469& 0.481& 0.459& 0.778& 0.659& {\color{red}\textbf{0.459}}& 0.465&  0.521& 0.496 \\

~&720&{\color{blue}\underline{0.491}}&{\color{blue}\underline{0.477}}&{\color{red}\textbf{0.483}}&{\color{red}\textbf{0.473}}&0.503& 0.491& 0.500& 0.488& 0.653& 0.621& 0.594& 0.558& 0.521& 0.500& 0.519& 0.516& 0.836& 0.699& 0.506& 0.507&  0.514& 0.512 \\
\cmidrule{2-24}
~&Avg&0.447&{\color{blue}\underline{0.440}}&{\color{blue}\underline{0.446}}&{\color{red}\textbf{0.439}}&0.454& 0.447& 0.469& 0.454& 0.529& 0.522& 0.541& 0.507& 0.458& 0.450& 0.456& 0.452& 0.747& 0.647& {\color{red}\textbf{0.440}}& 0.460&  0.496& 0.487 \\
\midrule
\multirow{5}*{\rotatebox{90}{ETTh2}}&96&{\color{red}\textbf{0.295}}&{\color{red}\textbf{0.348}}&0.298&0.350&{\color{blue}\underline{0.297}}& {\color{blue}\underline{0.349}}&  0.302& 0.348& 0.745& 0.584& 0.400& 0.440& 0.340& 0.374& 0.333& 0.387& 0.707& 0.621& 0.358& 0.397&  0.346& 0.388 \\

~&192&{\color{red}\textbf{0.380}}&{\color{red}\textbf{0.398}}&0.380&{\color{blue}\underline{0.399}}&0.380&0.400& {\color{blue}\underline{0.388}}& 0.400& 0.877& 0.656& 0.528& 0.509& 0.402& 0.414& 0.477& 0.476& 0.860& 0.689& 0.429& 0.439&  0.456& 0.452 \\

~&336&{\color{red}\textbf{0.420}}&{\color{red}\textbf{0.431}}&0.420&{\color{blue}\underline{0.432}}&0.428& 0.432&  {\color{blue}\underline{0.426}}& 0.433& 1.043& 0.731& 0.643& 0.571& 0.452& 0.452& 0.594& 0.541& 1.000& 0.744& 0.496& 0.487&  0.482& 0.486 \\

~&720&{\color{red}\textbf{0.419}}&{\color{red}\textbf{0.441}}&{\color{blue}\underline{0.425}}&{\color{blue}\underline{0.445}}&0.427& 0.445&  0.431& 0.446& 1.104& 0.763& 0.874& 0.679& 0.462& 0.468& 0.831& 0.657& 1.249& 0.838& 0.463& 0.474&  0.515& 0.511 \\
\cmidrule{2-24}
~&Avg&{\color{red}\textbf{0.379}}&{\color{red}\textbf{0.405}}&{\color{blue}\underline{0.381}}&{\color{blue}\underline{0.407}}& 0.383& 0.407&  0.387& 0.407& 0.942& 0.684& 0.611& 0.550& 0.414& 0.427& 0.559& 0.515& 0.954& 0.723& 0.437& 0.449&  0.450& 0.459 \\
\midrule
\multirow{5}*{\rotatebox{90}{ECL}}&96&{\color{blue}\underline{0.137}}&{\color{blue}\underline{0.232}}&{\color{red}\textbf{0.134}}&	{\color{red}\textbf{0.231}}&0.148& 0.240&  0.195& 0.285& 0.219& 0.314& 0.237& 0.329& 0.168& 0.272& 0.197& 0.282& 0.247& 0.345& 0.193& 0.308&  0.201& 0.317 \\

~&192&{\color{blue}\underline{0.155}}&{\color{red}\textbf{0.247}}&{\color{red}\textbf{0.152}}&{\color{blue}\underline{0.248}}& 0.162& 0.253&  0.199& 0.289& 0.231& 0.322& 0.236& 0.330& 0.184& 0.289& 0.196& 0.285& 0.257& 0.355& 0.201& 0.315&  0.222& 0.334 \\

~&336&{\color{blue}\underline{0.171}}&{\color{blue}\underline{0.265}}&{\color{red}\textbf{0.167}}&{\color{red}\textbf{0.262}}&0.178& 0.269&  0.215& 0.305& 0.246& 0.337& 0.249& 0.344& 0.198& 0.300& 0.209& 0.301& 0.269& 0.369& 0.214& 0.329&  0.231& 0.338 \\

~&720&{\color{blue}\underline{0.200}}&{\color{blue}\underline{0.290}}&{\color{red}\textbf{0.194}}&{\color{red}\textbf{0.288}}&0.225& 0.317&  0.256& 0.337& 0.280& 0.363& 0.284& 0.373& 0.220& 0.320& 0.245& 0.333& 0.299& 0.390& 0.246& 0.355&  0.254& 0.361 \\
\cmidrule{2-24}
~&Avg&{\color{blue}\underline{0.166}}&{\color{blue}\underline{0.259}}&{\color{red}\textbf{0.162}}&{\color{red}\textbf{0.257}}&0.178& 0.270&  0.216& 0.304& 0.244& 0.334& 0.251& 0.344& 0.192& 0.295& 0.212& 0.300& 0.268& 0.365& 0.214& 0.327&  0.227& 0.338 \\
\midrule
\multirow{5}*{\rotatebox{90}{Exchange}}&96&{\color{red}\textbf{0.085}}&{\color{red}\textbf{0.204}}&0.085&0.204&{\color{blue}\underline{0.086}}& 0.206&  0.088& {\color{blue}\underline{0.205}}& 0.256& 0.367& 0.094& 0.218& 0.107& 0.234& 0.088& 0.218& 0.267& 0.396& 0.148& 0.278&  0.197& 0.323 \\

~&192&{\color{red}\textbf{0.175}}&{\color{red}\textbf{0.299}}&0.177&{\color{blue}\underline{0.300}}&0.177& 0.299&  {\color{blue}\underline{0.176}}& 0.299& 0.470& 0.509& 0.184& 0.307& 0.226& 0.344& 0.176& 0.315& 0.351& 0.459& 0.271& 0.315&  0.300& 0.369 \\

~&336&0.330&{\color{blue}\underline{0.417}}&0.333&0.419&0.331& 0.417&  {\color{red}\textbf{0.301}}& {\color{red}\textbf{0.397}}& 1.268& 0.883& 0.349& 0.431& 0.367& 0.448& {\color{blue}\underline{0.313}}& 0.427& 1.324& 0.853& 0.460& 0.427&  0.509& 0.524 \\

~&720&{\color{blue}\underline{0.844}}&{\color{blue}\underline{0.695}}&0.857&0.699&0.847& {\color{red}\textbf{0.691}}&  0.901& 0.714& 1.767& 1.068& 0.852& 0.698& 0.964& 0.746& {\color{red}\textbf{0.839}}& 0.695& 1.058& 0.797& 1.195& {\color{blue}\underline{0.695}}&  1.447& 0.941 \\
\cmidrule{2-24}
~&Avg&{\color{blue}\underline{0.359}}&{\color{blue}\underline{0.404}}&0.363&0.406&0.360& {\color{red}\textbf{0.403}}&  0.367& 0.404& 0.940& 0.707& 0.370& 0.413& 0.416& 0.443& {\color{red}\textbf{0.354}}& 0.414& 0.750& 0.626& 0.519& 0.429&  0.613& 0.539 \\
\midrule
\multirow{5}*{\rotatebox{90}{Traffic}}&96&{\color{blue}\underline{0.401}}&{\color{blue}\underline{0.270}}&0.437&0.270&{\color{red}\textbf{0.395}}& {\color{red}\textbf{0.268}}&  0.544& 0.359& 0.522& 0.290& 0.805& 0.493& 0.593& 0.321& 0.650& 0.396& 0.788& 0.499& 0.587& 0.366&  0.613& 0.388 \\

~&192&{\color{blue}\underline{0.424}}&{\color{blue}\underline{0.281}}&0.462&0.282&{\color{red}\textbf{0.417}}& {\color{red}\textbf{0.276}}&  0.540& 0.354& 0.530& 0.293& 0.756& 0.474& 0.617& 0.336& 0.598& 0.370& 0.789& 0.505& 0.604& 0.373&  0.616& 0.382 \\

~&336&{\color{blue}\underline{0.445}}&{\color{blue}\underline{0.288}}&0.479&0.290&{\color{red}\textbf{0.433}}& {\color{red}\textbf{0.283}}&  0.551& 0.358& 0.558& 0.305& 0.762& 0.477& 0.629& 0.336& 0.605& 0.373& 0.797& 0.508& 0.621& 0.383&  0.622& 0.337 \\

~&720&{\color{blue}\underline{0.479}}&{\color{blue}\underline{0.306}}&0.531&0.322&{\color{red}\textbf{0.467}}& {\color{red}\textbf{0.302}}&  0.586& 0.375& 0.589& 0.328& 0.719& 0.449& 0.640& 0.350& 0.645& 0.394& 0.841& 0.523& 0.626& 0.382&  0.660& 0.408 \\
\cmidrule{2-24}
~&Avg&{\color{blue}\underline{0.437}}&{\color{blue}\underline{0.286}}&0.477&0.291&{\color{red}\textbf{0.428}}& {\color{red}\textbf{0.282}}&  0.555& 0.362& 0.550& 0.304& 0.760& 0.473& 0.620& 0.336& 0.625& 0.383& 0.804& 0.509& 0.610& 0.376&  0.628& 0.379 \\
\midrule
\multirow{5}*{\rotatebox{90}{Weather}}&96&{\color{blue}\underline{0.161}}&{\color{red}\textbf{0.206}}&0.165&{\color{blue}\underline{0.208}}&0.174& 0.214&  0.177& 0.218& {\color{red}\textbf{0.158}}& 0.230& 0.202& 0.261& 0.172& 0.220& 0.196& 0.255& 0.221& 0.306& 0.217& 0.296&  0.266& 0.336 \\

~&192&{\color{blue}\underline{0.208}}&{\color{red}\textbf{0.250}}&0.211&{\color{blue}\underline{0.251}}&0.221& 0.254&  0.225& 0.259& {\color{red}\textbf{0.206}}& 0.277& 0.242& 0.298& 0.219& 0.261& 0.237& 0.296& 0.261& 0.340& 0.276& 0.336&  0.307& 0.367 \\

~&336&{\color{red}\textbf{0.264}}&{\color{red}\textbf{0.291}}&{\color{blue}\underline{0.268}}&{\color{blue}\underline{0.293}}&0.278& 0.296&  0.278& 0.297& 0.272& 0.335& 0.287& 0.335& 0.280& 0.306& 0.283& 0.335& 0.309& 0.378& 0.339& 0.380&  0.359& 0.395 \\

~&720&{\color{blue}\underline{0.347}}&{\color{red}\textbf{0.346}}&0.350&{\color{blue}\underline{0.348}}&0.358& 0.349&  0.354& 0.348& 0.398& 0.418& 0.351& 0.386& 0.365& 0.359& {\color{red}\textbf{0.345}}& 0.381& 0.377& 0.427& 0.403& 0.428&  0.419& 0.428 \\
\cmidrule{2-24}
~&Avg&{\color{red}\textbf{0.245}}&{\color{red}\textbf{0.273}}&{\color{blue}\underline{0.249}}&{\color{blue}\underline{0.275}}&0.258& 0.279&  0.259& 0.281& 0.259& 0.315& 0.271& 0.320& 0.259& 0.287& 0.265& 0.317& 0.292& 0.363& 0.309& 0.360&  0.338& 0.382 \\
\midrule
\multirow{5}*{\rotatebox{90}{Solar-Energy}}&96&{\color{blue}\underline{0.198}}&0.239&{\color{red}\textbf{0.196}}&{\color{red}\textbf{0.232}}&0.203& {\color{blue}\underline{0.237}}&  0.234& 0.286& 0.310& 0.331& 0.312& 0.399& 0.250& 0.292& 0.290& 0.378& 0.237& 0.344& 0.242& 0.342&  0.884& 0.711 \\

~&192&{\color{blue}\underline{0.228}}&{\color{blue}\underline{0.259}}&{\color{red}\textbf{0.226}}&{\color{red}\textbf{0.256}}&0.233& 0.261&  0.267& 0.310& 0.734& 0.725& 0.339& 0.416& 0.296& 0.318& 0.320& 0.398& 0.280& 0.380& 0.285& 0.380&  0.834& 0.692 \\

~&336&{\color{red}\textbf{0.244}}&{\color{blue}\underline{0.272}}&{\color{blue}\underline{0.247}}&{\color{red}\textbf{0.271}}&0.248& 0.273&  0.290& 0.315& 0.750& 0.735& 0.368& 0.430& 0.319& 0.330& 0.353& 0.415& 0.304& 0.389& 0.282& 0.376&  0.941& 0.723 \\

~&720&{\color{red}\textbf{0.246}}&{\color{red}\textbf{0.275}}&0.250&0.275&{\color{blue}\underline{0.249}}& 0.275&  0.289& {\color{blue}\underline{0.317}}& 0.769& 0.765& 0.370& 0.425& 0.338& 0.337& 0.356& 0.413& 0.308& 0.388& 0.357& 0.427&  0.882& 0.717 \\
\cmidrule{2-24}
~&Avg&{\color{red}\textbf{0.229}}&{\color{blue}\underline{0.261}}&{\color{blue}\underline{0.230}}&{\color{red}\textbf{0.259}}&0.233& 0.262&  0.270& 0.307& 0.641& 0.639& 0.347& 0.417& 0.301& 0.319& 0.330& 0.401& 0.282& 0.375& 0.291& 0.381&  0.885& 0.711 \\

\bottomrule
\end{tabular}
}
\end{table*}

\begin{table*}
\centering
\caption{Full forecasting results of SOatten. The fixed input sequence length $L=96$ and the prediction lengths $T= \left\{ 96,192,336,720 \right\}$, which is consistent with the experimental setting in iTransformer \cite{liu2023itransformer}. \textbf{Bolded} results are superior to conventional attention. Avg means the average results of all prediction lengths.}
\label{Full SOatten Result}
\begin{tabular}{c|c|cccc|cccc}
\toprule
\multicolumn{2}{c|}{\multirow{2}*{Models}} 
&\multicolumn{4}{c|}{Temporal Transformer}&\multicolumn{4}{c}{Variate Transformer}\\

\multicolumn{2}{c|}{~} &\multicolumn{2}{c}{\textbf{SOatten(T)}}&\multicolumn{2}{c|}{PatchTST}&\multicolumn{2}{c}{\textbf{SOatten(V)}}&\multicolumn{2}{c}{iTransformer}\\
\cmidrule{3-10}
\multicolumn{2}{c|}{Metric}&MSE&MAE&MSE&MAE&MSE&MAE&MSE&MAE\\
\midrule

\multirow{5}*{\rotatebox{90}{ETTm1}}&96&\textbf{0.317}& \textbf{0.356}&0.329& 0.367& \textbf{0.329}& \textbf{0.365}&0.334& 0.368\\

~&192&\textbf{0.362}&\textbf{0.382}&0.367& 0.385&\textbf{0.370}& \textbf{0.387}&0.377& 0.391\\

~&336&\textbf{0.389}&\textbf{0.403}&0.399& 0.410&\textbf{0.401}&\textbf{0.407}&0.426& 0.420 \\

~&720&\textbf{0.451}&0.439&0.454& 0.439&\textbf{0.474}&\textbf{0.447}&0.491& 0.459  \\
\cmidrule{2-10}
~&Avg&\textbf{0.380}&\textbf{0.395}&0.387& 0.400&\textbf{0.394}&\textbf{0.402}&0.407& 0.410  \\
\midrule

\multirow{5}*{\rotatebox{90}{ETTm2}}&96&0.175&\textbf{0.258}&0.175& 0.259&0.180&0.264&0.180& 0.264\\

~&192&0.242&0.303&0.241& 0.302&\textbf{0.245}&\textbf{0.306}&0.250& 0.309\\

~&336&0.307&0.345&0.305& 0.343&0.312&0.349&0.311& 0.348 \\

~&720&\textbf{0.396}&\textbf{0.397}&0.402& 0.400&\textbf{0.411}&\textbf{0.406}&0.402& 0.400  \\
\cmidrule{2-10}
~&Avg&\textbf{0.280}&0.326&0.281& 0.326&\textbf{0.287}&\textbf{0.331}&0.288& 0.332   \\
\midrule

\multirow{5}*{\rotatebox{90}{ETTh1}}&96&\textbf{0.392}&\textbf{0.407}&0.414& 0.419&\textbf{0.383}&\textbf{0.400}&0.386& 0.405\\

~&192&\textbf{0.448}&\textbf{0.437}&0.460& 0.445&\textbf{0.440}&\textbf{0.433}&0.441& 0.436\\

~&336&\textbf{0.484}&\textbf{0.452}&0.501& 0.466&\textbf{0.475}&\textbf{0.449}&0.487& 0.458 \\

~&720&\textbf{0.479}&
\textbf{0.469}&0.500& 0.488&\textbf{0.491}&\textbf{0.477}&0.503& 0.491  \\
\cmidrule{2-10}
~&Avg&\textbf{0.451}&\textbf{0.441}&0.469& 0.454&\textbf{0.447}&\textbf{0.440}&0.454& 0.447  \\
\midrule

\multirow{5}*{\rotatebox{90}{ETTh2}}&96&\textbf{0.294}&\textbf{0.343}& 0.302& 0.348&\textbf{0.295}&\textbf{0.348}&0.297& 0.349 \\

~&192&\textbf{0.377}&\textbf{0.393}&0.388& 0.400&\textbf{0.380}&\textbf{0.398}&0.380& 0.400\\

~&336&\textbf{0.380}&\textbf{0.409}&0.426& 0.433&\textbf{0.420}&\textbf{0.431}&0.428& 0.432 \\

~&720&\textbf{0.412}&\textbf{0.433}&0.431& 0.446&\textbf{0.419}&\textbf{0.441}&0.427& 0.445   \\
\cmidrule{2-10}
~&Avg&\textbf{0.366}&\textbf{0.395}&0.387& 0.407&\textbf{0.379}&\textbf{0.405}&0.383& 0.407  \\
\midrule

\multirow{5}*{\rotatebox{90}{ECL}}&96&\textbf{0.176}&\textbf{0.260}&0.195&0.285&\textbf{0.137}& \textbf{0.232}&  0.148 &0.240\\

~&192&\textbf{0.182}&\textbf{0.267}& 0.199 &0.289&\textbf{0.155}& \textbf{0.247}&  0.162& 0.253\\

~&336&\textbf{0.198}&\textbf{0.283}&0.215 &0.305&\textbf{0.171}& \textbf{0.265}&  0.178& 0.269  \\

~&720&\textbf{0.239}&\textbf{0.316}&0.256& 0.337&\textbf{0.200}& \textbf{0.290}&  0.225 &0.317  \\
\cmidrule{2-10}
~&Avg&\textbf{0.199}&\textbf{0.282}&0.216 &0.304&\textbf{0.166}&\textbf{0.259}&  0.178 &0.270  \\
\midrule

\multirow{5}*{\rotatebox{90}{Exchange}}&96&\textbf{0.082}&\textbf{0.198}&0.088& 0.205 &\textbf{0.085}& \textbf{0.204}&  0.086 &0.206\\

~&192&\textbf{0.170}&\textbf{0.293}&0.176& 0.299&\textbf{0.175}& 0.299&  0.177 &0.299\\

~&336&0.314&0.403&0.301& 0.397&\textbf{0.330}& 0.417&  0.331 &0.417 \\

~&720&\textbf{0.872}&\textbf{0.699}&0.901 &0.714&\textbf{0.844}& 0.695&  0.847 &0.691   \\
\cmidrule{2-10}
~&Avg&\textbf{0.360}&\textbf{0.398}&0.367 &0.404&\textbf{0.359}& 0.404& 0.360 &0.403  \\
\midrule

\multirow{5}*{\rotatebox{90}{Traffic}}&96&\textbf{0.477}&\textbf{0.305}&0.544 &0.359&0.401& 0.270&  0.395 &0.268\\

~&192&\textbf{0.477}&\textbf{0.302}&0.540& 0.354&0.424& 0.281&  0.417& 0.276\\

~&336&\textbf{0.491}&\textbf{0.307}&0.551 &0.358&0.445& 0.288&  0.433 &0.283  \\

~&720&\textbf{0.524}&\textbf{0.326}&0.586& 0.375&0.479& 0.306&  0.467& 0.302   \\
\cmidrule{2-10}
~&Avg&\textbf{0.492}&\textbf{0.310}&0.555& 0.362&0.437& 0.286&  0.428 &0.282   \\
\midrule

\multirow{5}*{\rotatebox{90}{Weather}}&96&0.176&0.217&0.177 &0.218&\textbf{0.161}& \textbf{0.206}&  0.174 &0.214\\

~&192&0.223&0.258& 0.225 &0.259&\textbf{0.208}& \textbf{0.250}&  0.221 &0.254\\

~&336&0.276&0.298&0.278& 0.297&\textbf{0.264}& \textbf{0.291}&  0.278 &0.296 \\

~&720&\textbf{0.350}&\textbf{0.347}& 0.354 &0.348&\textbf{0.347}& \textbf{0.346}&  0.358& 0.349  \\
\cmidrule{2-10}
~&Avg&\textbf{0.256}&\textbf{0.280}&0.259& 0.281&\textbf{0.245}&\textbf{0.273}&  0.258 &0.279   \\
\midrule

\multirow{5}*{\rotatebox{90}{Solar-Energy}}&96&\textbf{0.227}&\textbf{0.264}&0.234 &0.286&\textbf{0.198}& 0.239&  0.203 &0.237 \\

~&192&\textbf{0.257}&\textbf{0.283}&0.267& 0.310&\textbf{0.228}& \textbf{0.259}&  0.233& 0.261 \\

~&336&\textbf{0.276}&\textbf{0.295}&0.290& 0.315&\textbf{0.244}& \textbf{0.272}&   0.248& 0.273 \\

~&720&\textbf{0.275}&\textbf{0.293}&0.289 &0.317&\textbf{0.246}& 0.275&   0.249& 0.275 \\
\cmidrule{2-10}
~&Avg&\textbf{0.259}&\textbf{0.284}&0.270 &0.307&\textbf{0.229}&\textbf{0.261}&  0.233& 0.262  \\

\bottomrule
\end{tabular}

\end{table*}

\subsection{Ablations}
\label{c2}

Table  \ref{Full MSS Result} shows the complete results of ablation for MSS. It can be seen that replacing MSS with a Linear map significantly reduced the forecasting performance of FSatten and SOatten, indicating that MSS is an essential design. Secondly, Table \ref{Full HCC Result} shows the complete results of ablation for HCC. The HCC module is very important for the orthogonal space learning and accurate generation of attention weights in SOatten.

\begin{table*}
\centering
\caption{Full results of ablation for MSS. The fixed input sequence length $L=96$ and the prediction lengths $T= \left\{ 96,192,336,720 \right\}$, which is consistent with the experimental setting in iTransformer \cite{liu2023itransformer}. Avg means the average results of all prediction lengths.}
\label{Full MSS Result}
\resizebox{\textwidth}{!}{
\begin{tabular}{c|c|cccc|cccc|cccc}
\toprule
\multicolumn{2}{c|}{\multirow{2}*{Models}} 
&\multicolumn{4}{c|}{Temporal Transformer}&\multicolumn{8}{c}{Variate Transformer}\\

\multicolumn{2}{c|}{~} &\multicolumn{2}{c}{SOatten(\textbf{L})}&\multicolumn{2}{c|}{SOatten}&\multicolumn{2}{c}{FSatten(\textbf{L})}&\multicolumn{2}{c}{FSatten}&\multicolumn{2}{c}{SOatten(\textbf{L})}&\multicolumn{2}{c}{SOatten}\\
\cmidrule{3-14}
\multicolumn{2}{c|}{Metric}&MSE&MAE&MSE&MAE&MSE&MAE&MSE&MAE&MSE&MAE&MSE&MAE\\
\midrule

\multirow{5}*{\rotatebox{90}{ETTm1}}&96&0.337&0.365&0.317&0.356&0.349&0.380&0.331&0.369&0.336&0.372&0.329&0.365\\

~&192&0.364&0.382&0.362&0.382&0.381&0.396&0.373&0.391&0.385&0.394&0.370&0.387\\

~&336&0.394&0.405&0.389&0.403&0.408&0.408&0.404&0.411&0.415&0.415&0.401&0.407 \\

~&720&0.457&0.440&0.451&0.439&0.471&0.445&0.469&0.447&0.483&0.454&0.474&0.447  \\
\cmidrule{2-14}
~&Avg&0.388&0.398&0.380&0.395&0.402&0.407&0.394&0.405&0.405&0.409&0.394&0.402  \\
\midrule

\multirow{5}*{\rotatebox{90}{ETTm2}}&96&0.180&0.264&0.175&0.258&0.185&0.269&0.180&0.265&0.182&0.267&0.180&0.264\\

~&192&0.243&0.304&0.242& 0.303&0.250&0.311&0.245&0.307&0.251&0.312&0.245& 0.306\\

~&336&0.302&0.341&0.307& 0.345&0.313&0.350&0.308&0.347&0.313&0.350&0.312& 0.349 \\

~&720&0.401&0.397&0.396& 0.397&0.410&0.403&0.412&0.405&0.411&0.404&0.411& 0.406  \\
\cmidrule{2-14}
~&Avg&0.282&0.326&0.280& 0.326&0.290&0.333&0.286&0.331&0.289&0.333&0.287& 0.331   \\
\midrule

\multirow{5}*{\rotatebox{90}{ETTh1}}&96&0.394&0.409&0.392& 0.407&0.383&0.400&0.384&0.401&0.383&0.401&0.383& 0.400\\

~&192&0.446&0.435&0.448& 0.437&0.436&0.430&0.434&0.429&0.435&0.431&0.440& 0.433\\

~&336&0.486&0.453&0.484& 0.452&0.479&0.451&0.481&0.453&0.475&0.451&0.475& 0.449 \\

~&720&0.481&0.471&0.479& 0.469&0.496&0.483&0.483&0.473&0.498&0.482&0.491& 0.477  \\
\cmidrule{2-14}
~&Avg&0.452&0.442&0.451& 0.441&0.449&0.441&0.446&0.439&0.448&0.441&0.447& 0.440  \\
\midrule

\multirow{5}*{\rotatebox{90}{ETTh2}}&96&0.295&0.343& 0.294& 0.343&0.305&0.354&0.298&0.350&0.304&0.352&0.295& 0.348 \\

~&192&0.377&0.393&0.377& 0.393&0.381&0.399&0.380&0.399&0.378&0.397&0.380& 0.398\\

~&336&0.382&0.409&0.380& 0.409&0.425&0.435&0.420&0.432&0.426&0.434&0.420& 0.431 \\

~&720&0.412&0.434&0.412& 0.433&0.431&0.449&0.425&0.445&0.421&0.441&0.419& 0.441   \\
\cmidrule{2-14}
~&Avg&0.367&0.395&0.366& 0.395&0.386&0.409&0.381&0.407&0.382&0.406&0.379& 0.405  \\
\midrule

\multirow{5}*{\rotatebox{90}{ECL}}&96&0.178&0.261&0.176&0.260&0.151& 0.242&0.134&0.231&0.149&0.241&  0.137&0.232\\

~&192&0.183&0.267&0.182&0.267&0.164& 0.256&0.152&0.248&0.163&0.254&0.155&0.247  \\

~&336&0.199&0.284&0.198&0.283&0.180& 0.272& 0.167&0.262&0.177&0.271&0.171&0.265\\

~&720&0.241&0 .317&0.239&0.316&0.212& 0.303& 0.194&0.288&0.212&0.301&0.200&0.290  \\
\cmidrule{2-14}
~&Avg&0.200& 0.282&0.199&0.282& 0.177&0.268&0.162&0.257&0.175&0.267&0.166&0.259  \\
\midrule

\multirow{5}*{\rotatebox{90}{Exchange}}&96&0.080&0.197&0.082&0.198&0.085& 0.204&0.085&0.204&0.085& 0.204& 0.085&0.204\\

~&192&0.169&0.292&0.170&0.293&0.177& 0.300&  0.177&0.300&0.177&0.299&0.175&0.299\\

~&336&0.335&0.417&0.314&0.403&0.333& 0.419&  0.333&0.419&0.335&0.417&0.330&0.417 \\

~&720&0.892&0.705&0.872&0.699&0.857&0.699&0.857&0.699&0.837&0.691&0.844&0.695  \\
\cmidrule{2-14}
~&Avg&0.369&0.403&0.360&0.398&0.363&0.406&0.363&0.406&0.359&0.403&0.359&0.404  \\
\midrule

\multirow{5}*{\rotatebox{90}{Traffic}}&96&0.477&0.305&0.477&0.305&0.439&0.270&0.437&0.270&0.404&0.275&0.401&0.270\\

~&192&0.475&0.302&0.477&0.302&0.464&0.282&0.462&0.282&0.422&0.281&0.424&0.281\\

~&336&0.489&0.308&0.491&0.307&0.480&0.290&0.479&0.290&0.448&0.289&0.445&0.288 \\

~&720&0.523&0.327&0.524&0.326&0.533&0.322&0.531&0.322&0.482&0.307&0.479&0.306  \\
\cmidrule{2-14}
~&Avg&0.491&0.311&0.492&0.310&0.479&0.291&0.477&0.291&0.439&0.288&0.437&0.286  \\
\midrule

\multirow{5}*{\rotatebox{90}{Weather}}&96&0.180&0.220&0.176&0.217&0.177&0.217&0.165&0.208&0.168&0.212&0.161&0.206\\

~&192&0.225&0.259&0.223&0.258&0.223&0.258&0.211&0.251&0.216&0.257&0.208&0.250\\

~&336&0.278&0.298&0.276&0.298&0.279&0.298&0.268&0.293&0.275&0.297&0.264&0.291 \\

~&720&0.350&0.347&0.350&0.347&0.356&0.349&0.350&0.348&0.353&0.347&0.347&0.346  \\
\cmidrule{2-14}
~&Avg&0.258&0.281&0.256&0.280&0.261&0.281&0.249&0.275&0.253&0.278&0.245&0.273 \\
\midrule

\multirow{5}*{\rotatebox{90}{Solar-Energy}}&96&0.226&0.265&0.227&0.264&0.199&0.237&0.196&0.232&0.207&0.241&0.198&0.239\\

~&192&0.255&0.284&0.257&0.283&0.229&0.262&0.226&0.256&0.235&0.262&0.228&0.259\\

~&336&0.275&0.295&0.276&0.295&0.250&0.276&0.247&0.271&0.249&0.275&0.244&0.272 \\

~&720&0.273&0.294&0.275&0.293&0.252&0.279&0.250&0.275&0.254&0.280&0.246&0.275  \\
\cmidrule{2-14}
~&Avg&0.257&0.285&0.259&0.284&0.233&0.264&0.230&0.259&0.236&0.265&0.229&0.261  \\

\bottomrule
\end{tabular}
}
\end{table*}

\begin{table*}
\centering
\caption{Full results of ablation for HCC. The fixed input sequence length $L=96$ and the prediction lengths $T= \left\{ 96,192,336,720 \right\}$, which is consistent with the experimental setting in iTransformer \cite{liu2023itransformer}. Avg means the average results of all prediction lengths.}
\label{Full HCC Result}
\begin{tabular}{c|c|cccc|cccc}
\toprule
\multicolumn{2}{c|}{\multirow{2}*{Models}} 
&\multicolumn{4}{c|}{Temporal Transformer}&\multicolumn{4}{c}{Variate Transformer}\\

\multicolumn{2}{c|}{~} &\multicolumn{2}{c}{\textbf{SOatten}}&\multicolumn{2}{c|}{w/o HCC}&\multicolumn{2}{c}{\textbf{SOatten}}&\multicolumn{2}{c}{w/o HCC}\\
\cmidrule{3-10}
\multicolumn{2}{c|}{Metric}&MSE&MAE&MSE&MAE&MSE&MAE&MSE&MAE\\
\midrule
\multirow{5}*{\rotatebox{90}{ETTm1}}&96&\textbf{0.317}& \textbf{0.356}&0.320& 0.361& \textbf{0.329}& \textbf{0.365}&0.336& 0.373\\

~&192&0.362&\textbf{0.382}&0.360& 0.383&\textbf{0.370}& \textbf{0.387}&0.375& 0.392\\

~&336&0.389&\textbf{0.403}&0.386& 0.404&\textbf{0.401}&\textbf{0.407}&0.407& 0.411 \\

~&720&0.451&0.439&0.447& 0.439&0.474&0.447&0.471& 0.447  \\
\cmidrule{2-10}
~&Avg&0.380&\textbf{0.395}&0.378& 0.397&\textbf{0.394}&\textbf{0.402}&0.397& 0.406  \\
\midrule

\multirow{5}*{\rotatebox{90}{ETTm2}}&96&\textbf{0.175}&\textbf{0.258}&0.178& 0.260&\textbf{0.180}&\textbf{0.264}&0.181& 0.266\\

~&192&\textbf{0.242}&\textbf{0.303}&0.243& 0.304&\textbf{0.245}&\textbf{0.306}&0.248& 0.309\\

~&336&\textbf{0.307}&0.345&0.308& 0.344&0.312&0.349&0.308& 0.346 \\

~&720&\textbf{0.396}&\textbf{0.397}&0.402& 0.401&\textbf{0.411}&0.406&0.412& 0.405  \\
\cmidrule{2-10}
~&Avg&\textbf{0.280}&\textbf{0.326}&0.283& 0.327&0.287&\textbf{0.331}&0.287& 0.332   \\
\midrule

\multirow{5}*{\rotatebox{90}{ETTh1}}&96&\textbf{0.392}&\textbf{0.407}&0.393& 0.409&\textbf{0.383}&0.400&0.384& 0.400\\

~&192&0.448&0.437&0.446& 0.436&0.440&0.433&0.437& 0.430\\

~&336&0.484&0.452&0.484& 0.452&\textbf{0.475}&\textbf{0.449}&0.480& 0.451 \\

~&720&0.479&0.469&0.477& 0.469&\textbf{0.491}&\textbf{0.477}&0.505& 0.489  \\
\cmidrule{2-10}
~&Avg&0.451&\textbf{0.441}&0.450& 0.442&\textbf{0.447}&\textbf{0.440}&0.451& 0.443  \\
\midrule

\multirow{5}*{\rotatebox{90}{ETTh2}}&96&\textbf{0.294}&0.343& 0.295& 0.343&0.295&0.348&0.295& 0.343 \\

~&192&0.377&0.393&0.377& 0.393&\textbf{0.380}&\textbf{0.398}&0.384& 0.402\\

~&336&\textbf{0.380}&0.409&0.381& 0.409&\textbf{0.420}&\textbf{0.431}&0.425& 0.435 \\

~&720&0.412&0.433&0.410& 0.432&\textbf{0.419}&\textbf{0.441}&0.430& 0.447   \\
\cmidrule{2-10}
~&Avg&0.366&0.395&0.366& 0.394&\textbf{0.379}&\textbf{0.405}&0.384& 0.407  \\
\midrule

\multirow{5}*{\rotatebox{90}{ECL}}&96&\textbf{0.176}&\textbf{0.260}&0.182&0.264&\textbf{0.137}& \textbf{0.232}&  0.138 &0.233\\

~&192&\textbf{0.182}&\textbf{0.267}& 0.187 &0.269&\textbf{0.155}& \textbf{0.247}&  0.158& 0.249\\

~&336&\textbf{0.198}&\textbf{0.283}&0.202 &0.285&\textbf{0.171}& \textbf{0.265}&  0.174& 0.269  \\

~&720&\textbf{0.239}&\textbf{0.316}&0.243& 0.319&\textbf{0.200}& \textbf{0.290}&  0.204 &0.293  \\
\cmidrule{2-10}
~&Avg&\textbf{0.199}&\textbf{0.282}&0.204 &0.284&\textbf{0.166}&\textbf{0.259}&  0.169 &0.261  \\
\midrule

\multirow{5}*{\rotatebox{90}{Exchange}}&96&0.082&\textbf{0.198}&0.082& 0.199 &\textbf{0.085}& \textbf{0.204}&  0.086 &0.205\\

~&192&\textbf{0.170}&\textbf{0.293}&0.175& 0.297&\textbf{0.175}& \textbf{0.299}&  0.176 &0.300\\

~&336&\textbf{0.314}&\textbf{0.403}&0.330& 0.414&\textbf{0.330}& \textbf{0.417}&  0.331 &0.418 \\

~&720&\textbf{0.872}&\textbf{0.699}&0.893 &0.707&\textbf{0.844}& 0.695&  0.846 &0.695   \\
\cmidrule{2-10}
~&Avg&\textbf{0.360}&\textbf{0.398}&0.370 &0.404&\textbf{0.359}& \textbf{0.404}& 0.360 &0.405  \\
\midrule

\multirow{5}*{\rotatebox{90}{Traffic}}&96&\textbf{0.477}&\textbf{0.305}&0.492 &0.315&\textbf{0.401}&\textbf{0.270}&  0.403 &0.272\\

~&192&\textbf{0.477}&\textbf{0.302}&0.495& 0.313&\textbf{0.424}& \textbf{0.281}&  0.426& 0.282\\

~&336&\textbf{0.491}&\textbf{0.307}&0.507 &0.318&\textbf{0.445}& \textbf{0.288}&  0.450 &0.291  \\

~&720&\textbf{0.524}&\textbf{0.326}&0.541& 0.336&0.479& \textbf{0.306}&  0.479& 0.307   \\
\cmidrule{2-10}
~&Avg&\textbf{0.492}&\textbf{0.310}&0.509& 0.321&\textbf{0.437}& \textbf{0.286}&  0.440 &0.288   \\
\midrule

\multirow{5}*{\rotatebox{90}{Weather}}&96&\textbf{0.176}&\textbf{0.217}&0.182 &0.222&\textbf{0.161}& \textbf{0.206}&  0.168 &0.211\\

~&192&\textbf{0.223}&\textbf{0.258}& 0.227 &0.260&\textbf{0.208}& \textbf{0.250}&  0.214 &0.252\\

~&336&\textbf{0.276}&0.298&0.279& 0.298&\textbf{0.264}& \textbf{0.291}&  0.270 &0.293 \\

~&720&\textbf{0.350}&0.347& 0.351 &0.346&\textbf{0.347}& 0.346&  0.349& 0.345  \\
\cmidrule{2-10}
~&Avg&\textbf{0.256}&\textbf{0.280}&0.260& 0.282&\textbf{0.245}&\textbf{0.273}&  0.250 &0.275   \\
\midrule

\multirow{5}*{\rotatebox{90}{Solar-Energy}}&96&\textbf{0.227}&\textbf{0.264}&0.230 &0.267&\textbf{0.198}& 0.239&  0.200 &0.238 \\

~&192&\textbf{0.257}&\textbf{0.283}&0.261& 0.286&\textbf{0.228}& \textbf{0.259}&  0.230& 0.261 \\

~&336&\textbf{0.276}&\textbf{0.295}&0.281& 0.298&\textbf{0.244}& \textbf{0.272}&   0.246& 0.273 \\

~&720&\textbf{0.275}&\textbf{0.293}&0.282 &0.297&\textbf{0.246}& \textbf{0.275}&   0.248& 0.276 \\
\cmidrule{2-10}
~&Avg&\textbf{0.259}&\textbf{0.284}&0.264 &0.287&\textbf{0.229}&\textbf{0.261}&  0.231& 0.262  \\

\bottomrule
\end{tabular} 

\end{table*}

\subsection{Efficiency}
\label{c3}

\begin{figure}[htbp]
    \centering
    \includegraphics[width=\textwidth]{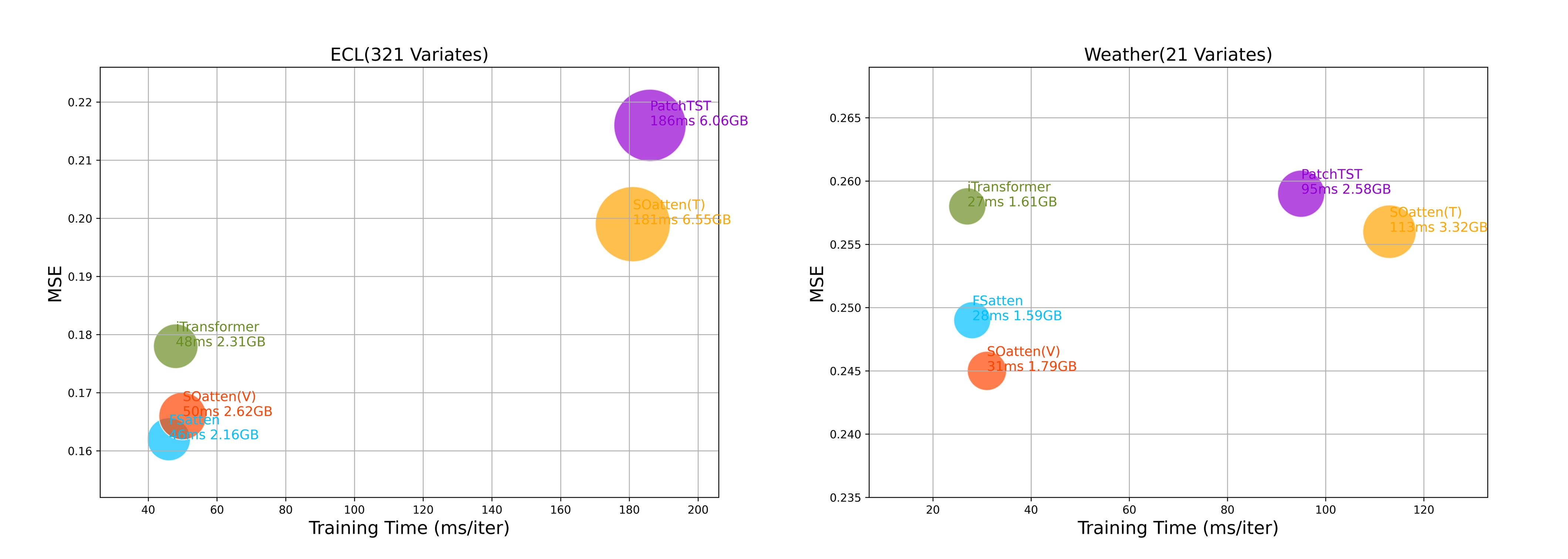}
    \caption{Model Efficiency of FSatten and SOatten compared with the conventional attention under input-96-predict-96 of Weather and ECL. A larger dot means a larger memory footprint.
}
    \label{Efficiency}
\end{figure}

Figure \ref{Efficiency} shows the efficiency results of FSatten and SOatten, compared with conventional attention under both the Temporal Transformer and the Variate Transformer. Since FSatten and SOatten were only replaced in both Transformers, with only the dimension size of MSS $F$ and an additional convolutional operation, there is no significant difference in efficiency compared to conventional attention. It can be observed that FSatten has slightly improved efficiency by replacing the original linear mapping of Query and Key with a Fast Fourier Transform \cite{brigham1967fast}. Secondly, because MSS is a Hadamard product, it can enhance the efficiency relative to fully connected layers to some extent. Although SOatten has an additional convolution layer, which significantly improves predictive performance, its memory usage and running time are slightly inferior to conventional attention.

\section{Showcases}
\label{d}

\begin{figure}[htbp]
    \centering
    \includegraphics[width=\textwidth]{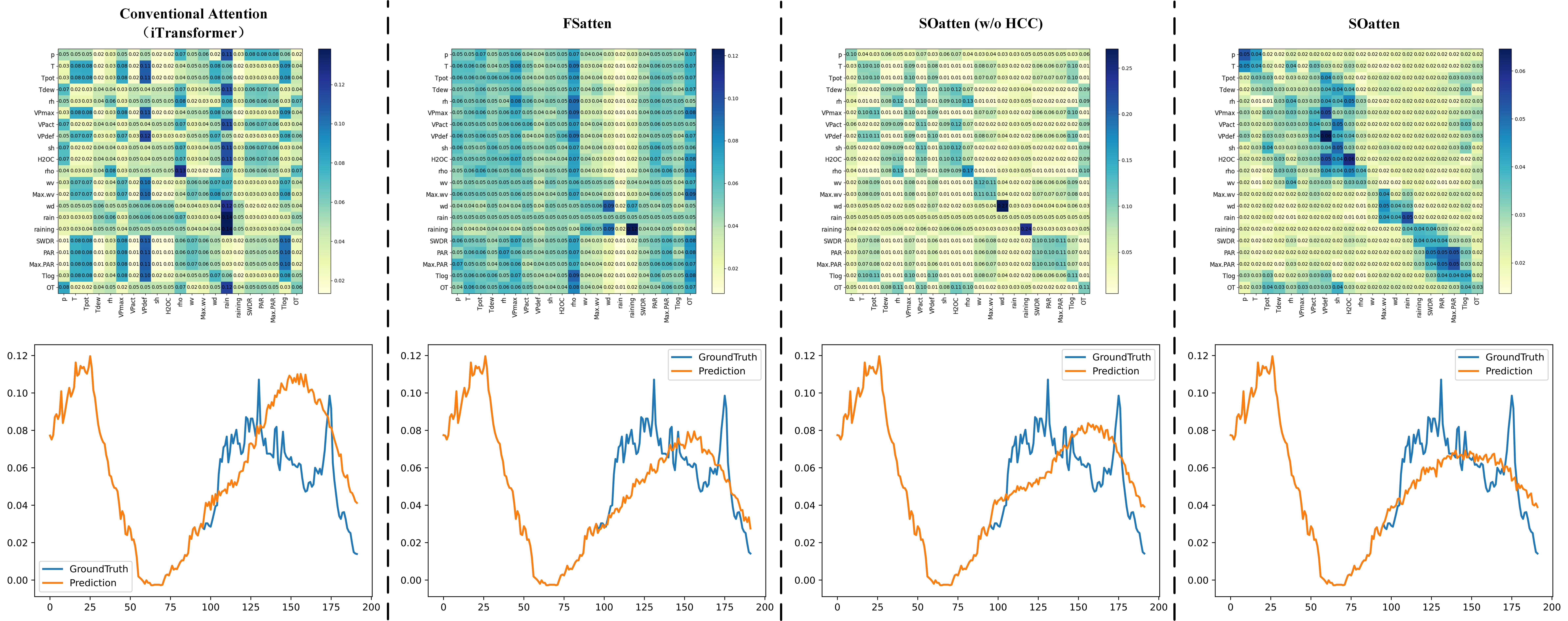}
    \caption{Attention maps and the forecasting of a few time series from Weather dataset under Variate Transformer. The attention map is calculated by averaging the attention matrices over all the heads and across all the layers.
}
    \label{Atten weather App}
\end{figure}

Figure \ref{Atten weather App} shows the attention weight values and corresponding predictions using various attention methods under the Variate Transformer. FSatten captures more accurate inter-variable frequency dependencies. SOatten learns association patterns based on physical characteristics in addition to sequence periodicity. Furthermore, by comparing the results of SOatten without HCC, it can be intuitively found that although it can learn various patterns, it cannot integrate multiple patterns accurately. The corresponding predictions are like a collage based on two patterns. This reversely proves the effectiveness of HCC. In addition, through numerical analysis of the attention weight matrix, the weight matrix of conventional attention is of low rank (19), while the FSatten and SOatten we proposed are full rank (21), showing the diversity when aggregating features from values. Moreover, by calculating the condition number, it is found that the condition numbers of the weight matrices generated by FSatten and SOatten are 1,519 and 1,480, respectively, which are much smaller than the conventional attention's condition number 78,596,560. This indicates that FSatten and SOatten can effectively capture more stable and reliable correlations between sequences than conventional attention. Condition Number Calculation Formula:
The condition number of a matrix $A$ is calculated as the ratio of the maximum singular value to the minimum singular value:
\begin{equation}
\kappa(A) = \frac{\sigma_{max}(A)}{\sigma_{min}(A)}
\end{equation}

\begin{figure}[htbp]
    \centering
    \includegraphics[width=\textwidth]{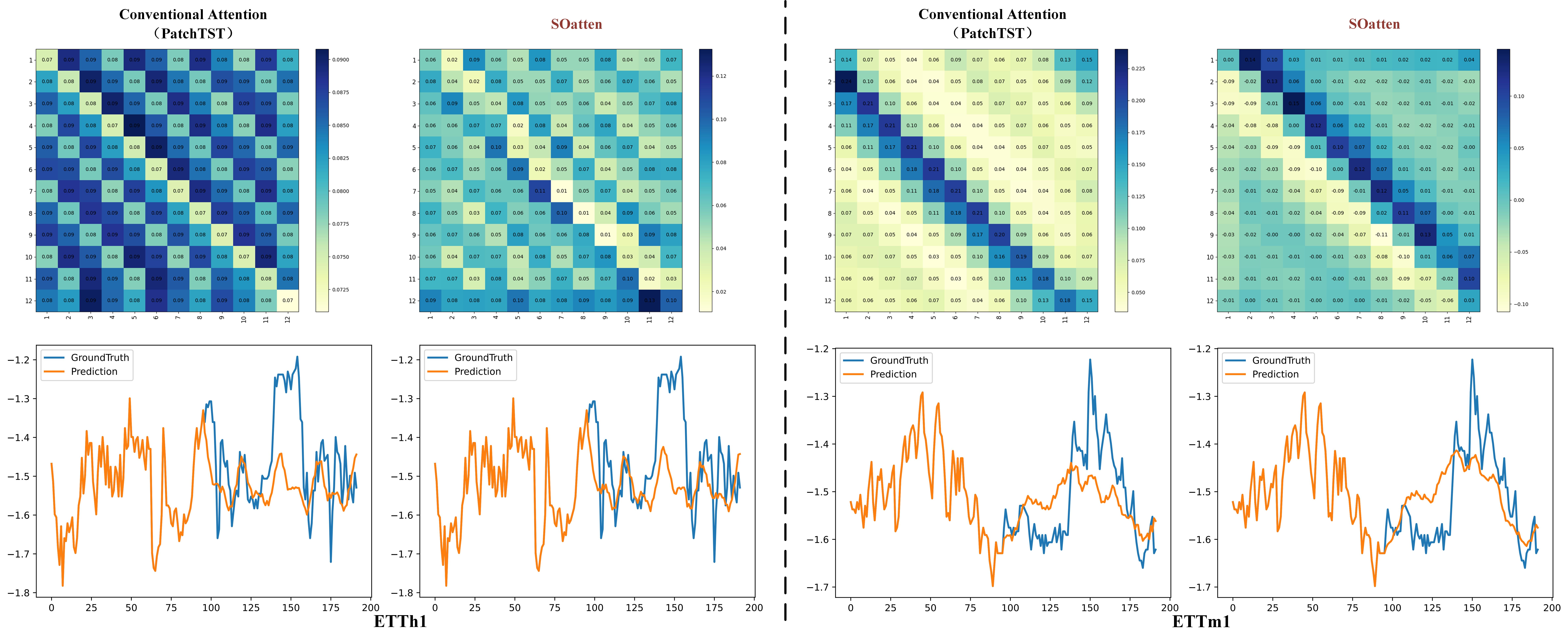}
    \caption{Attention maps and the forecasting of a few time series from ETT dataset under Temporal Transformer. The attention map is calculated by averaging the attention matrices over all the heads and across all the layers.
}
    \label{Atten Tem}
\end{figure}

Figure \ref{Atten Tem} demonstrates the attention weight values and corresponding forecasting results using SOatten under the Temporal Transformer. For the sub-sequences within the same variate, it can be observed that the attention weights generated by conventional attention are very uniform on ETTh1, and exhibit a very singular association on ETTm1. In contrast, SOatten is able to capture more comprehensive and precise dependency patterns compared to conventional attention.

\begin{figure}[htbp]
    \centering
    \includegraphics[width=\textwidth]{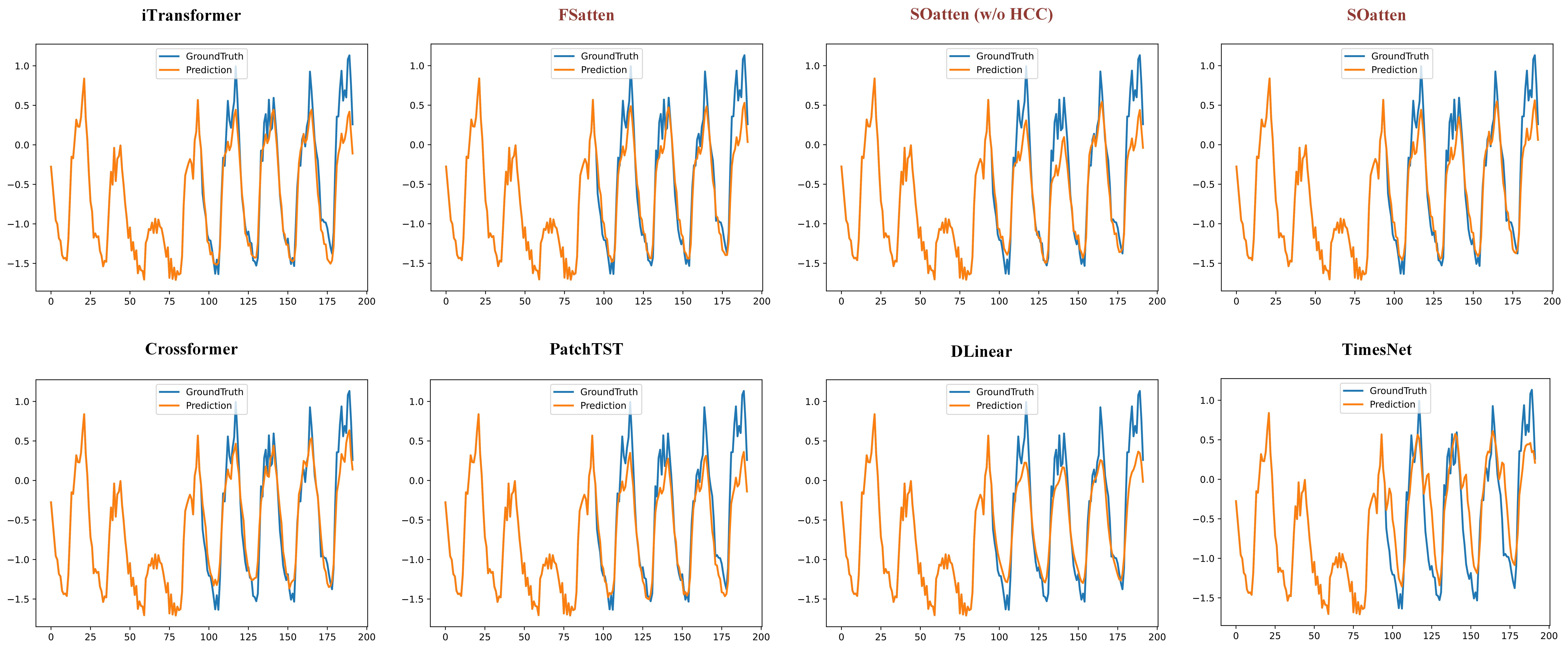}
    \caption{The forecasting of a few time series from the ECL dataset.}
    \label{Atten Tem}
\end{figure}

\end{appendix}

\begin{thebibliography}{10}

\bibitem{vaswani2017attention}
Ashish Vaswani, Noam Shazeer, Niki Parmar, Jakob Uszkoreit, Llion Jones, Aidan~N Gomez, {\L}ukasz Kaiser, and Illia Polosukhin.
\newblock Attention is all you need.
\newblock {\em Advances in neural information processing systems}, 30, 2017.

\bibitem{zhou2021informer}
Haoyi Zhou, Shanghang Zhang, Jieqi Peng, Shuai Zhang, Jianxin Li, Hui Xiong, and Wancai Zhang.
\newblock Informer: Beyond efficient transformer for long sequence time-series forecasting.
\newblock In {\em Proceedings of the AAAI conference on artificial intelligence}, volume~35, pages 11106--11115, 2021.

\bibitem{li2019enhancing}
Shiyang Li, Xiaoyong Jin, Yao Xuan, Xiyou Zhou, Wenhu Chen, Yu-Xiang Wang, and Xifeng Yan.
\newblock Enhancing the locality and breaking the memory bottleneck of transformer on time series forecasting.
\newblock {\em Advances in neural information processing systems}, 32, 2019.

\bibitem{liu2021pyraformer}
Shizhan Liu, Hang Yu, Cong Liao, Jianguo Li, Weiyao Lin, Alex~X Liu, and Schahram Dustdar.
\newblock Pyraformer: Low-complexity pyramidal attention for long-range time series modeling and forecasting.
\newblock In {\em International conference on learning representations}, 2021.

\bibitem{wu2021autoformer}
Haixu Wu, Jiehui Xu, Jianmin Wang, and Mingsheng Long.
\newblock Autoformer: Decomposition transformers with auto-correlation for long-term series forecasting.
\newblock {\em Advances in Neural Information Processing Systems}, 34:22419--22430, 2021.

\bibitem{zhou2022fedformer}
Tian Zhou, Ziqing Ma, Qingsong Wen, Xue Wang, Liang Sun, and Rong Jin.
\newblock Fedformer: Frequency enhanced decomposed transformer for long-term series forecasting.
\newblock In {\em International Conference on Machine Learning}, pages 27268--27286. PMLR, 2022.

\bibitem{nie2022time}
Yuqi Nie, Nam~H Nguyen, Phanwadee Sinthong, and Jayant Kalagnanam.
\newblock A time series is worth 64 words: Long-term forecasting with transformers.
\newblock {\em arXiv preprint arXiv:2211.14730}, 2022.

\bibitem{zhang2022crossformer}
Yunhao Zhang and Junchi Yan.
\newblock Crossformer: Transformer utilizing cross-dimension dependency for multivariate time series forecasting.
\newblock In {\em The Eleventh International Conference on Learning Representations}, 2022.

\bibitem{liu2023itransformer}
Yong Liu, Tengge Hu, Haoran Zhang, Haixu Wu, Shiyu Wang, Lintao Ma, and Mingsheng Long.
\newblock itransformer: Inverted transformers are effective for time series forecasting.
\newblock {\em arXiv preprint arXiv:2310.06625}, 2023.

\bibitem{zhou2023one}
Tian Zhou, Peisong Niu, Liang Sun, Rong Jin, et~al.
\newblock One fits all: Power general time series analysis by pretrained lm.
\newblock {\em Advances in neural information processing systems}, 36:43322--43355, 2023.

\bibitem{jin2023time}
Ming Jin, Shiyu Wang, Lintao Ma, Zhixuan Chu, James~Y Zhang, Xiaoming Shi, Pin-Yu Chen, Yuxuan Liang, Yuan-Fang Li, Shirui Pan, et~al.
\newblock Time-llm: Time series forecasting by reprogramming large language models.
\newblock {\em arXiv preprint arXiv:2310.01728}, 2023.

\bibitem{xu2023fits}
Zhijian Xu, Ailing Zeng, and Qiang Xu.
\newblock Fits: Modeling time series with $10 k $ parameters.
\newblock {\em arXiv preprint arXiv:2307.03756}, 2023.

\bibitem{brigham1967fast}
E~Oran Brigham and RE~Morrow.
\newblock The fast fourier transform.
\newblock {\em IEEE spectrum}, 4(12):63--70, 1967.

\bibitem{hornik1991approximation}
Kurt Hornik.
\newblock Approximation capabilities of multilayer feedforward networks.
\newblock {\em Neural networks}, 4(2):251--257, 1991.

\bibitem{yue2022ts2vec}
Zhihan Yue, Yujing Wang, Juanyong Duan, Tianmeng Yang, Congrui Huang, Yunhai Tong, and Bixiong Xu.
\newblock Ts2vec: Towards universal representation of time series.
\newblock In {\em Proceedings of the AAAI Conference on Artificial Intelligence}, volume~36, pages 8980--8987, 2022.

\bibitem{kiyasseh2021clocs}
Dani Kiyasseh, Tingting Zhu, and David~A Clifton.
\newblock Clocs: Contrastive learning of cardiac signals across space, time, and patients.
\newblock In {\em International Conference on Machine Learning}, pages 5606--5615. PMLR, 2021.

\bibitem{yeche2021neighborhood}
Hugo Y{\`e}che, Gideon Dresdner, Francesco Locatello, Matthias H{\"u}ser, and Gunnar R{\"a}tsch.
\newblock Neighborhood contrastive learning applied to online patient monitoring.
\newblock In {\em International Conference on Machine Learning}, pages 11964--11974. PMLR, 2021.

\bibitem{tonekaboni2021unsupervised}
Sana Tonekaboni, Danny Eytan, and Anna Goldenberg.
\newblock Unsupervised representation learning for time series with temporal neighborhood coding.
\newblock {\em arXiv preprint arXiv:2106.00750}, 2021.

\bibitem{lai2018modeling}
Guokun Lai, Wei-Cheng Chang, Yiming Yang, and Hanxiao Liu.
\newblock Modeling long-and short-term temporal patterns with deep neural networks.
\newblock In {\em The 41st international ACM SIGIR conference on research \& development in information retrieval}, pages 95--104, 2018.

\bibitem{zeng2023transformers}
Ailing Zeng, Muxi Chen, Lei Zhang, and Qiang Xu.
\newblock Are transformers effective for time series forecasting?
\newblock In {\em Proceedings of the AAAI conference on artificial intelligence}, volume~37, pages 11121--11128, 2023.

\bibitem{das2023long}
Abhimanyu Das, Weihao Kong, Andrew Leach, Rajat Sen, and Rose Yu.
\newblock Long-term forecasting with tide: Time-series dense encoder.
\newblock {\em arXiv preprint arXiv:2304.08424}, 2023.

\bibitem{liu2022scinet}
Minhao Liu, Ailing Zeng, Muxi Chen, Zhijian Xu, Qiuxia Lai, Lingna Ma, and Qiang Xu.
\newblock Scinet: Time series modeling and forecasting with sample convolution and interaction.
\newblock {\em Advances in Neural Information Processing Systems}, 35:5816--5828, 2022.

\bibitem{wu2022timesnet}
Haixu Wu, Tengge Hu, Yong Liu, Hang Zhou, Jianmin Wang, and Mingsheng Long.
\newblock Timesnet: Temporal 2d-variation modeling for general time series analysis.
\newblock In {\em The eleventh international conference on learning representations}, 2022.

\bibitem{wang2024timexer}
Yuxuan Wang, Haixu Wu, Jiaxiang Dong, Yong Liu, Yunzhong Qiu, Haoran Zhang, Jianmin Wang, and Mingsheng Long.
\newblock Timexer: Empowering transformers for time series forecasting with exogenous variables.
\newblock {\em arXiv preprint arXiv:2402.19072}, 2024.

\bibitem{gu2023mamba}
Albert Gu and Tri Dao.
\newblock Mamba: Linear-time sequence modeling with selective state spaces.
\newblock {\em arXiv preprint arXiv:2312.00752}, 2023.

\bibitem{ahamed2024timemachine}
Md~Atik Ahamed and Qiang Cheng.
\newblock Timemachine: A time series is worth 4 mambas for long-term forecasting.
\newblock {\em arXiv preprint arXiv:2403.09898}, 2024.

\bibitem{patro2024simba}
Badri~N Patro and Vijay~S Agneeswaran.
\newblock Simba: Simplified mamba-based architecture for vision and multivariate time series.
\newblock {\em arXiv preprint arXiv:2403.15360}, 2024.

\bibitem{sun2024learning}
Yu~Sun, Xinhao Li, Karan Dalal, Jiarui Xu, Arjun Vikram, Genghan Zhang, Yann Dubois, Xinlei Chen, Xiaolong Wang, Sanmi Koyejo, et~al.
\newblock Learning to (learn at test time): Rnns with expressive hidden states.
\newblock {\em arXiv preprint arXiv:2407.04620}, 2024.

\bibitem{kitaev2020reformer}
Nikita Kitaev, {\L}ukasz Kaiser, and Anselm Levskaya.
\newblock Reformer: The efficient transformer.
\newblock {\em arXiv preprint arXiv:2001.04451}, 2020.

\bibitem{dosovitskiy2020image}
Alexey Dosovitskiy, Lucas Beyer, Alexander Kolesnikov, Dirk Weissenborn, Xiaohua Zhai, Thomas Unterthiner, Mostafa Dehghani, Matthias Minderer, Georg Heigold, Sylvain Gelly, et~al.
\newblock An image is worth 16x16 words: Transformers for image recognition at scale.
\newblock {\em arXiv preprint arXiv:2010.11929}, 2020.

\bibitem{paszke2019pytorch}
Adam Paszke, Sam Gross, Francisco Massa, Adam Lerer, James Bradbury, Gregory Chanan, Trevor Killeen, Zeming Lin, Natalia Gimelshein, Luca Antiga, et~al.
\newblock Pytorch: An imperative style, high-performance deep learning library.
\newblock {\em Advances in neural information processing systems}, 32, 2019.

\end{thebibliography}
\end{document}